\documentclass[10pt,twocolumn,letterpaper]{article}

\usepackage[pagenumbers]{cvpr} 

\usepackage{amsthm}
\usepackage{algorithmic}
\usepackage{algorithm}

\theoremstyle{plain}
\newtheorem{theorem}{Theorem}[section]

\theoremstyle{definition}

\theoremstyle{remark}

\usepackage{ulem}
\usepackage{mathtools}
\usepackage{adjustbox}

\usepackage{xspace}
\newcommand{\algname}{\textit{EraseDiff}\xspace}

\usepackage{amsmath,amsfonts,bm}

\def\eqref#1{equation~\ref{#1}}

\def\1{\bm{1}}

\def\rvx{{\mathbf{x}}}
\def\rvy{{\mathbf{y}}}
\def\rvz{{\mathbf{z}}}

\def\vepsilon{{\bm{\epsilon}}}
\def\vSigma{{\bm{\Sigma}}}

\def\vzero{{\bm{0}}}

\def\vmu{{\bm{\mu}}}
\def\vtheta{{\bm{\theta}}}
\def\va{{\bm{a}}}

\def\vg{{\bm{g}}}

\def\mA{{\bm{A}}}

\DeclareMathAlphabet{\mathsfit}{\encodingdefault}{\sfdefault}{m}{sl}
\SetMathAlphabet{\mathsfit}{bold}{\encodingdefault}{\sfdefault}{bx}{n}

\def\gC{{\mathcal{C}}}
\def\gD{{\mathcal{D}}}

\def\gL{{\mathcal{L}}}

\def\gN{{\mathcal{N}}}
\def\gO{{\mathcal{O}}}

\def\gU{{\mathcal{U}}}

\def\sR{{\mathbb{R}}}

\newcommand{\E}{\mathbb{E}}

\DeclareMathOperator*{\argmin}{arg\,min}

\definecolor{cvprblue}{rgb}{0.21,0.49,0.74}
\usepackage[pagebackref,breaklinks,colorlinks,allcolors=cvprblue]{hyperref}

\title{Erasing Undesirable Influence in Diffusion Models}

\author{Jing Wu\textsuperscript{\rm 1}, Trung Le\textsuperscript{\rm 1}, Munawar Hayat\textsuperscript{\rm 2}, Mehrtash Harandi\textsuperscript{\rm 1}\\
\textsuperscript{\rm 1}Monash University, Melbourne, VIC, Australia, \textsuperscript{\rm 2}Qualcomm, San Deigo, CA, US\\
{\tt\small \{jing.wu1, trunglm, mehrtash.harandi\}@monash.edu, hayat@qti.qualcomm.com}
}

\begin{document}

\twocolumn[{%
\renewcommand\twocolumn[1][]{#1}%
\maketitle
\begin{center}
    \centering
    \captionsetup{type=figure} \includegraphics[width=\textwidth,keepaspectratio=True]{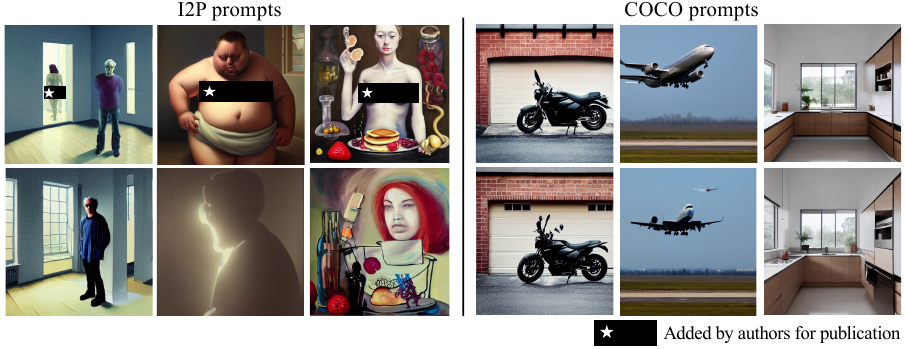}
    \captionof{figure}{Top to Bottom: generated samples by SD v1.4 and model scrubbed by our method, when erasing the concept of `nudity'. Our method can avoid NSFW (not safe for work) content while preserving model utility.}
\end{center}%
}]

\begin{abstract}
Diffusion models are highly effective at generating high-quality images but pose risks, such as the unintentional generation of NSFW (not safe for work) content.
Although various techniques have been proposed to mitigate unwanted influences in diffusion models while preserving overall performance, achieving a balance between these goals remains challenging.
In this work, we introduce EraseDiff, an algorithm designed to preserve the utility of the diffusion model on retained data while removing the unwanted information associated with the data to be forgotten.
Our approach formulates this task as a constrained optimization problem using the value function, resulting in a natural first-order algorithm for solving the optimization problem.
By altering the generative process to deviate away from the ground-truth denoising trajectory, we update parameters for preservation while controlling constraint reduction to ensure effective erasure, striking an optimal trade-off.
Extensive experiments and thorough comparisons with state-of-the-art algorithms demonstrate that EraseDiff effectively preserves the model's utility, efficacy, and efficiency.

{\color{red}WARNING:} This paper contains sexually explicit imagery that may be offensive in nature.

\end{abstract}

\section{Introduction}
\label{sec:intro}

Diffusion Models~\cite{ho2020denoising,song2020denoising,rombach2021high} are now the method of choice in deep generative models, owing to their high-quality output, stability, and ease of training procedure. This has facilitated their successful integration into commercial applications such as \textit{midjourney}.
Unfortunately, the ease of use associated with diffusion models brings forth significant privacy risks. 
Studies have shown that these models can memorize and regenerate individual images from their training datasets~\cite{somepalli2023diffusion,somepalli2023understanding,carlini2023extracting}.
Beyond privacy, diffusion models are susceptible to misuse and can generate NSFW digital content~\cite{rando2022red,salman2023raising,schramowski2023safe}. 
In this context, individuals whose images are used for training might request the removal of their private data. 
In particular, data protection regulations like the European Union General Data Protection Regulation (GDPR)~\cite{voigt2017eu} and the California Consumer Privacy Act (CCPA)~\cite{goldman2020introduction} grant users the \textit{right to be forgotten}, obligating companies to expunge data pertaining to a user upon receiving a request for deletion.
These legal provisions grant data owners the right to remove their data from trained models and eliminate its influence on said models~\cite{bourtoule2021machine,guo2019certified,golatkar2020eternal,mehta2022deep,sekhari2021remember,ye2022learning,tarun2023deep,tarun2023fast,chen2023boundary}.

A straightforward solution is to retrain the model from scratch after excluding the data that needs to be forgotten. However, the removal of pertinent data followed by retraining diffusion models from scratch demands substantial resources and is often deemed impractical.
A version of the stable diffusion model trained on subsets of the LAION-5B dataset~\cite{schuhmann2022laion} costs approximately 150,000 GPU hours with 256 A100 GPUs\footnote{https://stablediffusion.gitbook.io/overview/stable-diffusion-overview/technology/training-procedures}.
Existing research on erasing unwanted influence has primarily focused on classification tasks~\cite{karasuyama2010multiple, cao2015towards,ginart2019making,bourtoule2021machine, wu2020deltagrad,guo2019certified,golatkar2020eternal,mehta2022deep,sekhari2021remember,chen2023boundary}.
Despite substantial progress, prior methods developed in classification are observed to be ineffective for generation tasks~\cite{fan2023salun}.
Consequently, there is a pressing need for the development of methods capable of scrubbing data from diffusion models without necessitating complete retraining.

Recently, a handful of studies~\cite{gandikota2023erasing,gandikota2023unified,zhang2023forget,heng2023continual,heng2023selective,kumari2023ablating,fan2023salun,lyu2024one,bui2024erasing} target unlearning in diffusion models, with a primary focus on the text-to-image models~\cite{gandikota2023erasing,gandikota2023unified,zhang2023forget}.
Broadly, these methods aim to achieve two main objectives: erasing data influence and preserving overall model performance.
However, as demonstrated by \citet{bui2024removing}, balancing this trade-off remains challenging.

In this work, we propose \algname, an algorithm tailored to balance the overall performance of diffusion models with the erasure of undesirable information.
Drawing inspiration from optimization-based meta-learning algorithms~\cite{maclaurin2015gradient,finn2017model} that enable fast adaptation to new learning tasks, we formulate this challenge as a bi-level optimization problem, where the ``inner" optimization focus on erasing undesirable influence and the ``outer" objective seeks to preserve model performance.
The outer objective and inner optimization are interdependent, iterating between preservation and erasure to balance the trade-off effectively.
However, this nested optimization can be challenging to optimize efficiently. The inner optimization may converge to a saddle point or struggle with non-convex functions, making it difficult to achieve a stable solution~\cite{liu2022bome}. 
Therefore, we further reformulate the problem as a constrained optimization problem using the value function~\cite{liu2021value,outrata1990numerical,ye1995optimality}, which facilitates a natural first-order solution~\cite{liu2022bome}, allows us to optimize preservation and erasure in a unified manner.  
This approach achieves a fine-tuned balance between preservation and targeted erasure, yielding an optimal trade-off.
We benchmark \algname on various scenarios, encompassing unlearning of classes on CIFAR-10~\cite{krizhevsky2009learning} with Denoising Diffusion Probabilistic Models (DDPM)~\cite{ho2020denoising}, classes on Imagenette~\cite{howard2020fastai} and concepts on the I2P dataset~\cite{schramowski2023safe} with stable diffusion.
Our empirical findings show that \algname is $11\times$ faster than Heng and Soh's method~\cite{heng2023selective} and $2\times$ faster than Fan's method~\cite{fan2023salun} when forgetting on DDPM while achieving better unlearning results across several metrics.
The results demonstrate that \algname is capable of effectively erasing data influence in diffusion models, ranging from specific classes to the concept of nudity.

\section{Background}
\label{sec:back}


In this section, we outline the components of the models we evaluate, including DDPM and latent diffusion models~\citep{rombach2021high}.
Throughout the paper, we denote scalars, and vectors/matrices by lowercase and bold symbols, respectively (\eg, $a$, $\va$, $\mA$).

\paragraph{DDPM.}
(1) Diffusion: DDPM gradually diffuses the data distribution $ \sR^d \ni \rvx_0 \sim q(\rvx)$ into the standard Gaussian distribution $\sR^d \ni \vepsilon \sim \gN(\vzero, \mathbf{I}_d)$ with $T$ time steps, \ie, $q(\rvx_t|\rvx_{t-1}) = \gN(\rvx_t;\sqrt{\alpha_t}\rvx_{t-1}, (1-\alpha_t)\mathbf{I}_d)$, where $\alpha_t=1-\beta_t$ and $\{\beta_t\}^T_{t=1}$ are the pre-defined variance schedule. The diffusion takes the form $\rvx_t$ as $\rvx_t=\sqrt{\Bar{\alpha}_t}\rvx_0 + \sqrt{1 - \Bar{\alpha}_t}\vepsilon$, where $\Bar{\alpha}_t=\prod_{i=1}^t \alpha_i$.
(2) Training: A model $\epsilon_{\vtheta}(\cdot)$ with parameters $\vtheta \in \sR^n$ is trained to learn the reverse process $p_{\vtheta}(\rvx_{t-1}|\rvx_t) \approx q(\rvx_{t-1}|\rvx_t)$.
Given $\rvx_0 \sim q(\rvx)$ and time step $t \in [1, T]$, the simplified training objective is to minimize the distance between $\vepsilon$ and the predicted $\vepsilon_t$ given $\rvx_0$ at time $t$, \ie, $\| \vepsilon - \epsilon_{\vtheta}(\rvx_t, t)\|$.
(3) Sampling: after training the model, we could obtain the learnable backward distribution $p_{\vtheta^*}(\rvx_{t-1}|\rvx_t) = \gN(\rvx_{t-1}; \vmu_{\vtheta^*}(\rvx_t, t), \vSigma_{\vtheta^*}(\rvx_t, t))$, where $\vmu_{\vtheta^*}(\rvx_t,t)=\frac{1}{\sqrt{\alpha_t}}(\rvx_t - \frac{\beta_t}{\sqrt{1-\alpha_t}}\epsilon_{\vtheta}(\rvx_t, t))$
and $\vSigma_{\vtheta^*}(\rvx_t,t)=\frac{(1-\Bar{\alpha}_{t-1})\beta_t}{1-\Bar{\alpha}_t}$.
Then, given $\rvx_T \sim \gN(\vzero, \mathbf{I}_d)$, $\rvx_0$ could be obtained via sampling from $p_{\vtheta^*}(\rvx_{t-1}|\rvx_t)$ from $t=T$ to $t=1$ step by step.
%


\paragraph{Latent diffusion models.}
Latent diffusion models apply the diffusion models in the latent space $\rvz$ of a pre-trained variational autoencoder. The noise would be added to $\rvz=\varepsilon(\rvx)$, instead of the data $\rvx$, and the denoised output would be transformed to image space with the decoder.
Besides, text embeddings generated by models like CLIP are used as conditioning inputs.

\section{Diffusion Unlearning}
\label{sec:method}
Let $\gD=\{\rvx_i, c_i\}_i^N$ be a dataset of images $\rvx_i$ associated with label $c_i$ representing the class. $\gC=\{1, \cdots, C\}$ denotes the label space where $C$ is the total number of classes and $c_i \in \gC$.
We split the training data $\gD$ into the forgetting data $\gD_f \subset \gD$ and its complement, remaining data $\gD_r = \gD \setminus \gD_f$.
The forgetting data has label space $\gC_f \subseteq \gC$, 
and the remaining label space is denoted as $\gC_r= \gC \setminus \gC_f$.

\subsection{Training objective}
Our goal is to scrub the information about $\gD_f$ carried by the diffusion models while maintaining the model utility over the remaining data $\gD_r$.
To achieve this, we adopt different training objectives for $\gD_r$ and $\gD_f$ as follows.

For the remaining data $\gD_r$, we fine-tune the diffusion models with the original objective:
\begin{align}
    \label{eq:obj_dr}
    \gL_r(\vtheta; \gD_r) = \E_{t, \vepsilon \in \gN(\vzero, \mathbf{I}_d), (\rvx_0, c) \sim \gD_r \times \gC_r} [ \| \vepsilon - \epsilon_{\vtheta}(\rvx_t|c) \|_{2}^{2} ],
\end{align}
where $\rvx_t=\sqrt{\Bar{\alpha}_t}\rvx_0 + \sqrt{1 - \Bar{\alpha}_t}\vepsilon$.
For the forgetting data $\gD_f$, we aim to let the models fail to generate meaningful images corresponding to $\gC_f$ and thus propose:
\begin{align}
    \label{eq:obj_df}
    \gL_f(\vtheta; \gD_f) = \E_{t, \vepsilon \in \gN(\vzero, \mathbf{I}_d), (\rvx_0, c) \sim \gD_f \times \gC_f} [ \| \vepsilon_f - \epsilon_{\vtheta}(\rvx_t|c) \|_{2}^{2} ],
\end{align}
where $\vepsilon_f=\epsilon_{\vtheta}(\rvx_t|c_m)$ and $c_m \neq c$ so that the denoised image $\rvx_0$ is not related to the forgetting class/concept $c$~\cite{fan2023salun,heng2023selective}.
With this, we hinder the approximator $\epsilon_{\vtheta}$ to guide the denoising process to obtain meaningful examples for the forgetting data example $\rvx_0 \sim \gD_f$. 

To erase the undesirable influence of $\gD_f$ and preserve the overall performance, it is common to form 
\begin{align}
    \label{eq:obj_init}
    \gL_r(\vtheta; \gD_{r}) + \lambda \gL_f(\vtheta; \gD_{f}),
\end{align}
with $\lambda > 0$ as the optimization objective (see for example \cite{fan2023salun}).
However, training could be hindered due to the conflicting gradients between the erasing and preservation objectives, preventing a balanced trade-off between erasure and preservation~\cite{liu2021conflict}.
To address this, rather than scalarizing the two objectives, we consider a framework based on optimization-based meta-learning algorithm~\cite{rajeswaran2019meta} that allows iteratively updates to optimize each objective:
\begin{align}
    \label{eq:obj_final_mh}
    &\operatorname{min}_{\vtheta}\;\gL_r(\vtheta; \gD_{r}) \notag \\
    \text{s.t.} \quad &\vtheta \in \argmin_{\bm{\phi}} \gL_f(\bm{\phi} ; \gD_{f})\;,
\end{align}
where the outer objective minimizes the remaining loss $\gL_r$ (\ie, preserving model utility), the inner optimization minimizes the forgetting loss $\gL_f$ (\ie, erasing) with initialization $\mathbb{R}^n \ni \bm{\phi}_{\texttt{init}} = \vtheta$.
Given $\vtheta$, the inner optimization on $\bm{\phi}$ aims to minimize the forgetting data influence, with the goal of achieving effective erasure while preserving model utility. 
The outer objective and inner optimization are interdependent, iterating between preservation and erasure to balance the trade-off effectively.

While the above framework allows for iterative updates to address the conflicting objectives of erasure and preservation, it still relies on nested optimization, which can be challenging to optimize efficiently. The inner optimization may converge to a saddle point or struggle with non-convex functions, making it difficult to achieve a stable solution~\cite{liu2022bome}. 
To further streamline the optimization process, we adopt a value function approach~\cite{liu2021value,outrata1990numerical,ye1995optimality} that reformulates the problem as a single-constrained optimization:
\begin{align}
    \label{eq:obj_final}
    &\operatorname{min}_{\vtheta}\;\gL_r(\vtheta; \gD_{r}) \notag \\
    \text{s.t.} \quad & \gL_f(\vtheta; \gD_{f}) - \operatorname{min}_{\bm{\phi}}\gL_f(\bm{\phi}; \gD_{f}) \leq 0,
\end{align}
where $\bm{\phi}$ is initialized at $\vtheta$, leverages the value function to encapsulate the influence of data erasure directly as a constraint on $\gL_f$.
This avoids the need for a nested loop by capturing the forgetting objective as a constraint and provides a natural first-order solution, as the constrained formulation allows us to optimize preservation and erasure in a unified manner.

\subsection{Solution}
To solve \cref{eq:obj_final}, let us first denote $g(\vtheta) \mathrel{:}= \gL_f(\vtheta; \gD_{f})-\operatorname{min}_{\bm{\phi}} \gL_f(\bm{\phi};\gD_{f})$.
Our goal is to erase undesirable influence while preserving the overall model performance, hence the update vector $\bm{\delta}_{t}$ for updating the model should aid in minimizing $\gL_r(\vtheta;\gD_{r})$ and $g(\vtheta)$ simultaneously.
In other words, suppose that the current solution for \cref{eq:obj_final} is $\vtheta_{t}$, we aim to update $\vtheta_{t+1}=\vtheta_{t}-\eta \bm{\delta}_{t}$ where $\eta$ is sufficiently small, so that $\gL_r(\vtheta_{t+1};\gD_{r})$ decreases (\ie, preserve model utility) and $g(\vtheta_{t+1})$ decreases (\ie, erasure).
To this end, we aim to find the update vector $\bm{\delta}_{t}$ by:
\begin{align}
    \label{eq:delta_opt}
    &\bm{\delta}_{t} \in \frac{1}{2}\operatorname{argmin}_{\bm{\delta}} \big\| \nabla_{\vtheta}\gL_r(\vtheta_{t};\gD_{r})-\bm{\delta} \big \|_{2}^{2}, \notag \\
    \text{s.t.} \quad &\nabla_{\vtheta} g(\vtheta_{t})^{\top} \bm{\delta}\geq a_{t} > 0.
\end{align}
This will ensure that the update $\bm{\delta}_{t}$ is close to $\nabla_{\vtheta}\gL_r(\vtheta_{t};\gD_{r})$ and decreases $g(\vtheta_{t})$ until it reaches stationary.
Because $g\left(\vtheta_{t+1}\right)-g\left(\vtheta_{t}\right)\approx-\eta\nabla_{\vtheta}g\left(\vtheta_{t}\right)^{\top}\bm{\delta}\leq-\eta a_{t}<0$ for some scalar $a_{t} > 0$, we can ensure that $g\left(\vtheta_{t+1}\right)<g\left(\vtheta_{t}\right)$ for small step size $\eta>0$.
This means that the update $\bm{\delta}_{t}$ can ensure to minimize $\gL_f(\vtheta;\gD_f)$ as long as it does not conflict with descent of $\gL_r(\vtheta;\gD_r)$.

To find the solution to the optimization problem in \cref{eq:delta_opt}, the following theorem is developed:
\begin{theorem}
The optimal solution of the optimization problem in \cref{eq:delta_opt} is $\bm{\delta}^{*} = \nabla_{\vtheta} \gL_r(\vtheta_{t};\gD_{r}) + \lambda_{t} \nabla_{\vtheta} g(\vtheta_{t})$ where $\lambda_{t} = \operatorname{max} \{0, \frac{a_{t} - \nabla_{\vtheta}g(\vtheta_{t})^{\top} \nabla_{\vtheta} \gL_r(\vtheta_{t}; \gD_{r})} {\Vert\nabla_{\vtheta} g(\vtheta_{t})\Vert_{2}^{2}}\}$.
\end{theorem}

We provide the proof in \textsection\ref{sec:appendix_proof} in the Appendix.
This provides the solution to the optimization problem by constructing the update vector $\bm{\delta}$ to balance two competing objectives.
The variable $a_{t}$ adjusts the weight of the forgetting objective, ensuring that the update vector $\bm{\delta}$ decreases the remaining loss without violating the erasure goal, and achieves the dual goals of maintaining utility and achieving erasing.
In practice, we can choose $a_{t}=\eta \Vert\nabla_{\vtheta} g(\vtheta_{t})\Vert_{2}^{2}$, and we start from $\bm{\phi}^{0}=\vtheta_{t}$ and use gradient descend in $K$ steps with the learning rate $\xi>0$ to reach $\bm{\phi}^{K}$, namely $\bm{\phi}^{k+1} = \bm{\phi}^{k} - \xi \nabla_{\bm{\phi}} \gL_f(\bm{\phi}^{k};\gD_f)$ and $k=0, \cdots, K-1$.

\begin{algorithm}[tb]
\caption{\algname: Erasing undesirable influence in diffusion models.}
\label{alg: pseudocode}
\begin{algorithmic}[1]
    \REQUIRE Well-trained model with parameters $\vtheta_0$, forgetting data $\gD_f$ and remaining data $\gD_r$, outer iteration number $T$ and inner iteration number $K$, learning rate $\eta$.
    \ENSURE Parameters $\vtheta^*$ for the scrubbed model.
    \FOR{iteration $t$ in $T$}
        \STATE $\bm{\phi}^0 = \vtheta_t$.
        \STATE Get $\bm{\phi}^K$ by $K$ steps of gradient descent on $\gL_f(\bm{\phi}; \gD_f)$ starting from $\bm{\phi}^0$.
        \STATE Set $g(\vtheta_t) = \gL_f(\vtheta_t; \gD_f) - \gL_f(\bm{\phi}^K; \gD_f)$.
        \STATE Update the model: $\vtheta_{t+1} = \vtheta_t - \eta( \nabla_{\vtheta_t} \gL_r(\vtheta_t; \gD_r) + \lambda_t \nabla_{\vtheta_t} g(\vtheta_t; \bm{\phi}^K) )$, 
        \STATE where $\lambda_t =\operatorname{max} \{0,\frac{a_{t} - \nabla_{\vtheta} g(\vtheta_{t})^{T} \nabla_{\vtheta} \gL_r(\vtheta_{t};\gD_{r})} {\Vert\nabla_{\vtheta} g(\vtheta_{t}) \Vert_{2}^{2}}\}$.
    \ENDFOR 
\end{algorithmic}
\end{algorithm}

\subsection{Analysis}
We can characterize the solution of our algorithm as follows and the proof can be found in \textsection\ref{sec:appendix_proof} in the Appendix:
\begin{theorem}[Pareto optimality]
The stationary point obtained by our algorithm is Pareto optimal of the problem $\operatorname{min}_{\vtheta}[\gL_r(\vtheta;\gD_r), \gL_f(\vtheta;\gD_f)]$.
\end{theorem}
This asserts that the solution obtained by our algorithm is Pareto optimal for the problem of minimizing both objectives, which implies that the solution obtained by the algorithm ensures a balanced trade-off between preserving model utility and erasing undesirable influences.

\begin{figure}[tb]
  \centering
  \includegraphics[width=0.48\textwidth, keepaspectratio=True]{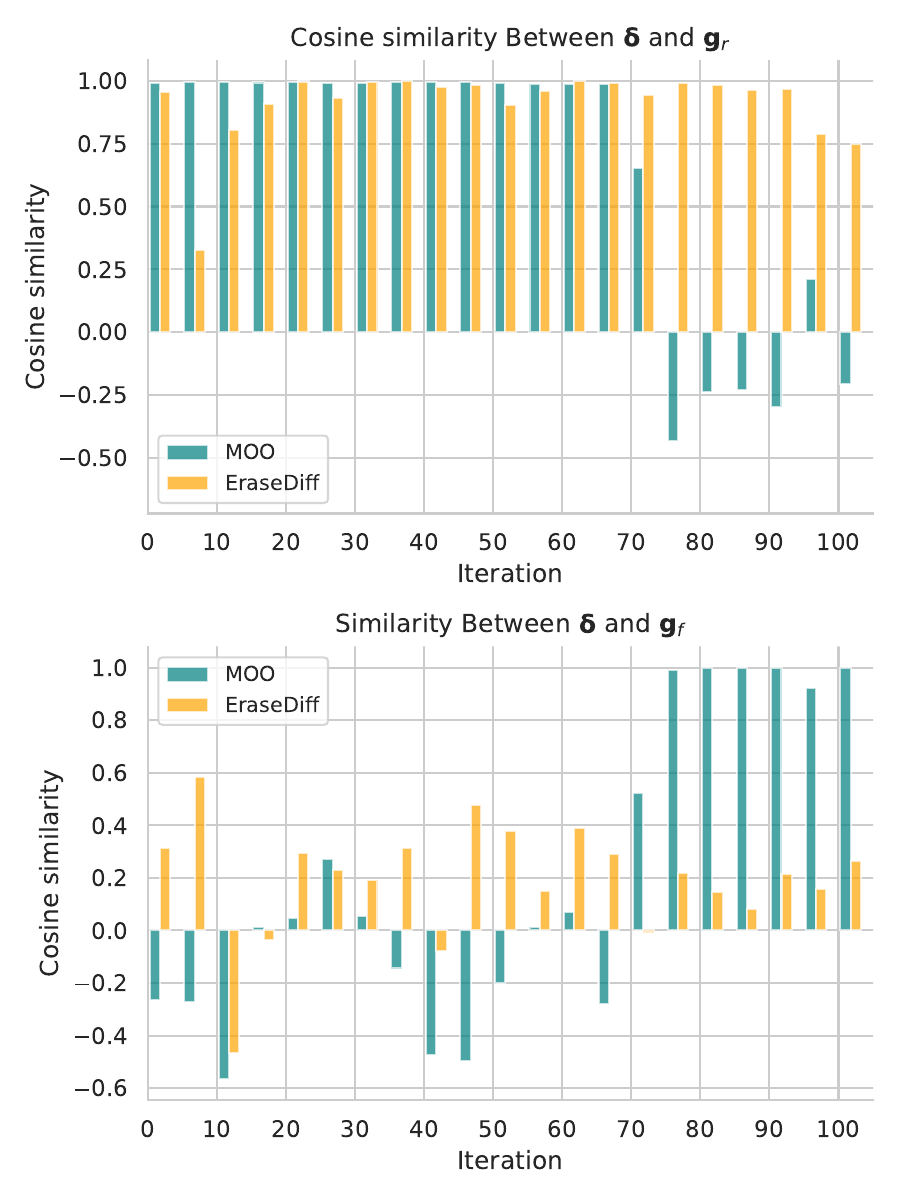}
  \caption{Top to bottom: cosine similarity between the update vector $\bm{\delta}$ and the preservation gradient $\vg_r$, followed by the cosine similarity between $\bm{\delta}$ and the erasing gradient $\vg_f$. Positive values indicate alignment, while negative values suggest conflict. This visualization illustrates how well the update vector aligns with the objectives of preservation and erasure over successive iterations.}
  \label{fig:similarity}
\end{figure}

We further take DDPM with CIFAR-10 when forgetting the `airplane' as an example to show that our proposed method helps alleviate the gradient conflict which prevents a balanced trade-off between erasure and preservation.
\cref{fig:similarity} presents the cosine similarity between the update vector $\bm{\delta}$ and the gradient $\vg_r=\nabla_{\vtheta} \gL_r(\vtheta; \gD_{r})$ for preservation, and the cosine similarity between the update vector $\bm{\delta}$ and the gradient $\vg_f=\nabla \gL_f(\vtheta; \gD_{f})$ for erasing.
%
The cosine similarity represents the alignment between the update vector and the gradients associated with the preservation and erasure. Higher positive values indicate alignment, meaning that the update vector $\bm{\delta}$ is directed similarly to the respective gradient, whereas negative values indicate misalignment or conflict. In particular, negative values suggest a high degree of opposition between the update vector and the respective gradient, which can signify competing objectives between preservation and erasure during the optimization process.

MOO (Multi-Objective Optimization) denotes the naive integration of erasing and preservation as stated in \cref{eq:obj_init}.
For the vanilla MOO, \cref{fig:similarity} shows a clear alternating pattern in cosine similarity values between the update vector $\bm{\delta}$ and the gradients $\vg_r$ and $\vg_f$. Specifically, when the cosine similarity between $\bm{\delta}$ and $\vg_r$ is greater than 0, the similarity between $\bm{\delta}$ and $\vg_f$ tends to be less than 0, and vice versa.
This pattern suggests that MOO experiences gradient conflict, as it cannot effectively balance the two objectives of preservation and erasure, preventing MOO from achieving a harmonious update that supports both goals simultaneously.
In contrast, \algname mostly shows positive cosine similarity values between the update vector $\bm{\delta}$ and both the preservation gradient $\vg_r$ and the erasing gradient $\vg_f$.
This indicates that \algname aligns the update direction with both objectives, suggesting it manages to avoid significant gradient conflict. By maintaining positive alignment, \algname appears to balance preservation and erasure more effectively, leading to better cooperation between objectives.

\section{Related Work}
\label{sec:relatedwork}

\paragraph{Memorization in generative models.}
Privacy of generative models has been studied extensively for GANs~\citep{feng2021gans,meehan2020non,webster2021person} and generative language models~\citep{carlini2022quantifying,carlini2021extracting,jagielski2022measuring,tirumala2022memorization}.
These generative models often risk replicating from their training data.
Recently, several studies~\citep{carlini2023extracting,somepalli2023understanding,somepalli2023diffusion,vyas2023provable} investigated these data replication behaviors in diffusion models, raising concerns about the privacy and copyright issues.
Possible mitigation strategies are deduplicating and randomizing conditional information~\citep{somepalli2023understanding,somepalli2023diffusion}, or training models with differential privacy (DP)~\citep{abadi2016deep,dwork2006calibrating,dwork2008differential,dockhorn2022differentially}.
However, leveraging DP-SGD~\citep{abadi2016deep} may cause training to diverge~\citep{carlini2023extracting}.

\paragraph{Malicious misuse.}
Diffusion models usually use training data from varied open sources and when such
unfiltered data is employed, there is a risk of it being tainted~\citep{chen2023trojdiff} or manipulated~\citep{rando2022red}, resulting in inappropriate generation~\citep{schramowski2023safe}.
They also risk the imitation of copyrighted content, \eg, mimicking the artistic style~\citep{gandikota2023erasing,shan2023glaze}.
To counter inappropriate generation, data censoring~\citep{gandhi2020scalable,birhane2021large,nichol2021glide,schramowski2022can} where excluding black-listed images before training, and safety guidance where diffusion models will be updated away from the inappropriate/undesired concept~\citep{gandikota2023erasing, schramowski2023safe} are proposed.
\citet{shan2023glaze} propose protecting artistic style by adding barely perceptible perturbations to the artworks before public release.
Yet, \citet{rando2022red} argue that DMs can still generate content that bypasses the filter.
\citet{chen2023trojdiff} highlight the susceptibility of DMs to poison attacks, where target images are generated with specific triggers.

\paragraph{Machine unlearning.}
Removing data directly involves retraining the model from scratch, which is inefficient and impractical.
Thus, to reduce the computational overhead, efficient machines unlearning methods~\citep{romero2007incremental,karasuyama2010multiple, cao2015towards,ginart2019making,bourtoule2021machine, wu2020deltagrad,guo2019certified,golatkar2020eternal,mehta2022deep,sekhari2021remember,chen2023boundary,tarun2023deep} have been proposed.
Several studies~\citep{gandikota2023erasing,gandikota2023unified,heng2023continual,heng2023selective,fan2023salun,zhang2023forget,bui2024removing} recently introduce unlearning in diffusion models.
Most of them~\citep{gandikota2023erasing, gandikota2023unified,heng2023continual,zhang2023forget} mainly focus on text-to-image models and high-level visual concept erasure.
\citet{heng2023selective} adopt Elastic Weight Consolidation (EWC) and Generative Replay (GR) from continual learning to perform unlearning effectively without access to the training data. Heng and Soh's method can be applied to a wide range of generative models, however, it needs the computation of FIM for different datasets and models, which may lead to significant computational demands.
\citet{fan2023salun} propose a very potent unlearning algorithm called SalUn that shifts attention to important parameters \wrt the forgetting data. SalUn can perform effectively across image classification and generation tasks.

In this work, we introduce a simple yet effective unlearning algorithm for diffusion models by formulating the problem as a constrained optimization problem, to achieve a fine-tuned balance between preservation and targeted erasure, yielding an optimal trade-off.
Below, we will show that our algorithm is not only faster than Heng and Soh's method~\citep{heng2023selective} and Fan's method~\citep{fan2023salun}, but even outperforms these methods in terms of the trade-off between the forgetting and preserving model utility.

\section{Experiment}
\label{sec:exp}
We evaluate \algname in various scenarios, including removing images with specific classes/concepts, to answer the following research questions (RQs):
(\romannumeral1) Can typical machine unlearning methods be applied to diffusion models?
(\romannumeral2) Is \algname able to remove the influence of $\gD_f$ in the diffusion models?
(\romannumeral3) Is \algname able to preserve the model utility while removing $\gD_f$?
(\romannumeral4) Is \algname efficient in removing the data?
(\romannumeral5) How does \algname perform on the public well-trained models?

\subsection{Setup}
Experiments are reported on CIFAR-10~\cite{krizhevsky2009learning} with DDPM, Imagenette~\cite{howard2020fastai} with Stable Diffusion (SD) for class-wise forgetting, I2P~\cite{schramowski2023safe} dataset with SD for concept-wise forgetting.
For all SD experiments, we use the open-source SD v1.4~\cite{rombach2021high} checkpoint as the pre-trained model.
Implementation details and additional results like visualizations of generated images can be found in \textsection\ref{sec:appendix_de} and \textsection\ref{sec:appendix_ar}.

\paragraph{Baselines.}
We primarily benchmark against the following baselines commonly used in machine unlearning:
(\romannumeral1) \textit{Unscrubbed}, (\romannumeral2) \textit{Finetune (FT)}~\cite{golatkar2020eternal}, (\romannumeral3) \textit{NegGrad (NG)}~\cite{golatkar2020eternal}, (\romannumeral4) \textit{BlindSpot}~\cite{tarun2023deep}, (\romannumeral5) \textit{ESD}~\cite{gandikota2023erasing}, (\romannumeral6) \textit{FMN}~\cite{zhang2023forget}, (\romannumeral7) \textit{Selective Amnesia (SA)}~\cite{heng2023selective} and (\romannumeral8) the SOTA machine unlearning algorithm  \textit{SalUn}~\cite{fan2023salun}.

\paragraph{Metrics.}
Several metrics are utilized to evaluate the algorithms:
(\romannumeral1) \textit{Frechet Inception Distance (FID)}~\cite{heusel2017gans}: the widely-used metric for assessing the quality of generated images.
(\romannumeral2) \textit{CLIP score}: the similarity between the visual features of the generated image and its corresponding textual embedding.
(\romannumeral3) $P_{\psi}(\rvy=c_f|\rvx_f)$~\cite{heng2023selective}: the classification rate of a pre-trained classifier $P_{\psi}(\rvy|\rvx)$, with a ResNet architecture~\cite{he2016deep} used to classify generated images conditioned on the forgetting classes. A lower classification value indicates superior unlearning performance.
(\romannumeral4) \textit{Precision and Recall}:
A low FID may indicate high precision (realistic images) but low recall (small variations)~\cite{sajjadi2018assessing,kynkaanniemi2019improved}.
\citet{kynkaanniemi2019improved} shows that generative models claim to optimize FID (high fidelity) but always sacrifice variation (low diversity).
Hence, we include metric precision (fidelity) and recall (diversity) to express the quality of the generated samples, to provide explicit visibility of the tradeoff between sample quality and variety.

\begin{table*}[tb]
    \centering
    \caption{Results on CIFAR10 with DDPM when forgetting the `airplane' class. $P_{\psi}(\rvy=c_f|\rvx_f)$ indicate the probability of the forgotten class (\ie, the effectiveness of erasing). Precision and Recall demonstrate the fidelity and diversity~\cite{sajjadi2018assessing,kynkaanniemi2019improved}, and FID scores are computed between the generated 45K images and the corresponding ground truth images with the same labels from $\gD_r$ (\ie, preserving model utility). SA excels in class-wise forgetting but struggles to perform concept-wise forgetting as shown in \cref{fig:i2p_change} and \cref{tab:nsfw_sd}.
    The best and the second best are highlighted in {\color{cyan}{blue}} and {\color{orange}{orange}}, respectively.}
    \label{tab:cifar10_ddpm}
    \begin{tabular}{l|cccccc|cc}
        \toprule
        &Unscrubbed &FT~\cite{golatkar2020eternal} &NG~\cite{golatkar2020eternal} &BlindSpot~\cite{tarun2023deep} &SA~\cite{heng2023selective} &SalUn~\cite{fan2023salun} &EraseDiff$_{\text{rl}}$ &EraseDiff$_{\text{noise}}$ \\ 
        \midrule
        FID $\downarrow$            &9.63 &{8.21} &76.73 &9.12 &\colorbox{orange!30}{8.19} &9.16 &8.66 &\colorbox{cyan!50}{\textbf{7.61}} \\ 
        Precision (fidelity) $\uparrow$        &0.40 &\colorbox{cyan!50}{\textbf{0.43}} &0.08  &0.41 &\colorbox{cyan!50}{\textbf{0.43}} &0.41 &\colorbox{cyan!50}{\textbf{0.43}} &\colorbox{cyan!50}{\textbf{0.43}} \\ 
        Recall (diversity) $\uparrow$           &0.79 &\colorbox{orange!30}{0.77} &0.61  &\colorbox{cyan!50}{\textbf{0.78}} &0.75 &0.76 &\colorbox{orange!30}{0.77} &0.72 \\ 
        $P_{\psi}(\rvy=c_f|\rvx_f)$$\downarrow$ &0.97 &0.96 &0.61  &0.90 &\colorbox{cyan!50}{\textbf{0.06}} &\colorbox{orange!30}{0.07} &0.24 &0.22\\ 
        \bottomrule
    \end{tabular}
\end{table*}
\begin{figure*}[tb]
  \centering
  \includegraphics[width=0.98\textwidth, keepaspectratio=True]{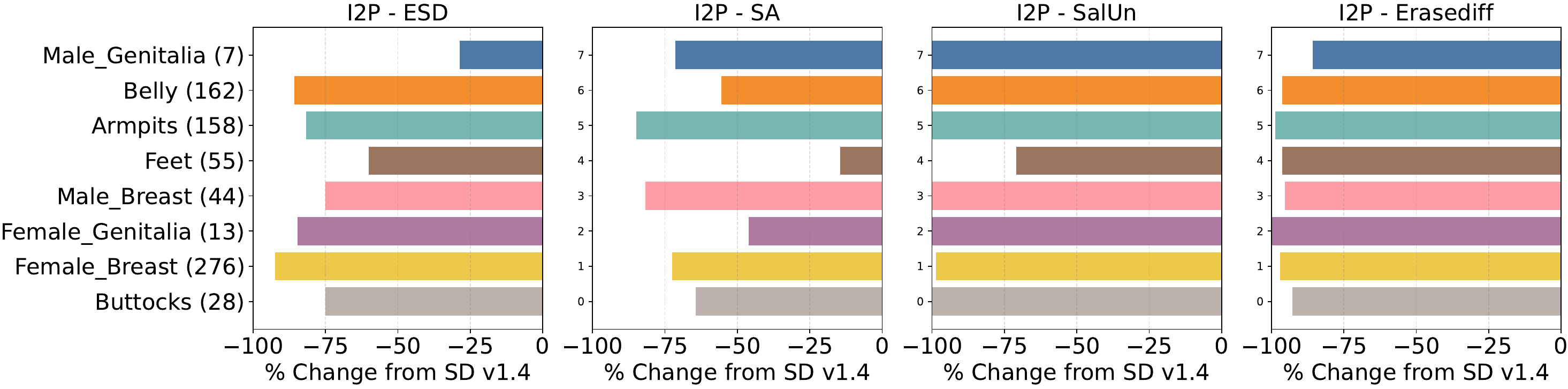}
  \caption{Quantity of nudity content detected using the NudeNet classifier from I2P data. Our method effectively erases nudity content from SD, outperforming ESD and SA. Note that \cref{fig:i2p_change} and \cref{tab:nsfw_sd} together presents the trade-off between erasing and preservation.}
  \label{fig:i2p_change}
\end{figure*}
\begin{table}[tb]
    \centering
    \caption{Evaluation of 30K generated images by SD when erasing `nudity'. The FID score is measured compared to validation data, while the CLIP similarity score evaluates the alignment between generated images and the corresponding prompts. The best and the second best are highlighted in {\color{cyan}{blue}} and {\color{orange}{orange}}, respectively.}
    \label{tab:nsfw_sd}
    \begin{tabular}{l|cccc}
        \toprule
        &ESD~\cite{gandikota2023erasing} &SA~\cite{heng2023selective} &SalUn~\cite{fan2023salun} &\algname \\ 
        \midrule
        FID $\downarrow$  &\colorbox{cyan!50}{\textbf{15.76}} &25.58 &25.06 &\colorbox{orange!30}{17.01}  \\ 
        CLIP $\uparrow$   &30.33 &\colorbox{cyan!50}{\textbf{31.03}} &28.91 &\colorbox{orange!30}{30.58}   \\ 
        \bottomrule
    \end{tabular}
\end{table}

\subsection{Results on DDPM}
Following SA, we aim to forget the `airplane' class on CIFAR-10.
Here,  we replace $\vepsilon \in \gN(\vzero, \mathbf{I}_d)$ with $\vepsilon_f = \epsilon_{\vtheta}(\rvx_t|c_m)$ like random labelling used in \cite{fan2023salun} where $c_m \neq c$, denoted as EraseDiff$_{\text{rl}}$. We also try to use $\vepsilon_f = \gU(\vzero, \mathbf{I}_d)$ like SA, denoted as EraseDiff$_{\text{noise}}$.
Note that the choice of replacement for forgotten classes is flexible and is not the primary focus of this work. For further discussion on the choice of substitution strategies, please refer to related studies~\cite{lyu2024one,bui2024erasing}.

Results are presented in \cref{tab:cifar10_ddpm}.
Firstly, from \cref{tab:cifar10_ddpm}, we can conclude that traditional machine unlearning methods designed for image classification or regression tasks fall short in effectively performing forgetting for DDPM. Finetune and BlindSpot suffer from under-forgetting (\ie, the generated image quality is good but the probability of generated images belonging to the forgetting class approaching the value of the unscrubbed model), and NegGrad suffers from over-forgetting (the probability of generated images belonging to the forgetting class is decreased compared to that of the unscrubbed model but the generated image quality drops significantly).

Then, comparing SA and SalUn's unlearning methods, SA achieves an FID score of 8.19 but sacrifices variation (decreased recall). Also, note that SA introduces excessive computational resource requirements and time consumption~\cite{heng2023selective,zhang2024unlearncanvas}.
Note that the FID scores of SA, SalUn, and \algname decrease compared with the generated images from the original models; the quality of the generated images experiences a slight improvement.
However, there is a decrease in recall (diversity), which can be attributed to the scrubbed models being fine-tuned over $\gD_r$, suggesting a tendency towards overfitting.
Regarding forgetting, SalUn achieves a smaller probability of the generated images classified as the forgetting class than ours; yet, the FID score is larger than ours, and images generated by EraseDiff$_{\text{rl}}$ present better diversity and fidelity. 

\subsection{Results on Stable Diffusion}
\begin{table*}[tb]
    \centering
    \caption{Performance of class-wise forgetting on Imagenette using SD. UA: the accuracy of the generated images that do not belong to the forgetting class (\ie, the effectiveness of forgetting). The FID score is measured compared to validation data for the remaining classes.}
    \label{tab:imagenette_sd}
    \begin{adjustbox}{max width=0.78\textwidth}
    \begin{tabular}{l|cc|cc|cc|cc}
        \toprule
        Forget. Class &\multicolumn{2}{c|}{FMN$^\ast$~\cite{zhang2023forget}} &\multicolumn{2}{c|}{ESD$^\ast$~\cite{gandikota2023erasing}} &\multicolumn{2}{c|}{SalUn$^\ast$~\cite{fan2023salun}} &\multicolumn{2}{c}{\algname} \\
        &FID $\downarrow$ &UA (\%)$\uparrow$ &FID $\downarrow$ &UA (\%)$\uparrow$ &FID $\downarrow$ &UA (\%)$\uparrow$ &FID $\downarrow$ &UA (\%)$\uparrow$  \\
        \midrule
         Tench            &1.63 &42.40 &1.22 &99.40   &2.53 &100.00  &1.29 &100 \\
         English Springer &1.75 &27.20 &1.02 &100.00  &0.79 &100.00  &1.38 &100 \\
         Cassette Player  &0.80 &93.80 &1.84 &100.00  &0.91 &99.80   &0.85 &100 \\
         Chain Saw        &0.94 &48.40 &1.48 &96.80   &1.58 &100.00  &1.17 &99.9 \\
         Church           &1.32 &23.80 &1.91 &98.60   &0.90 &99.60   &0.83 &100 \\
         French Horn      &0.99 &45.00 &1.08 &99.80   &0.94 &100.00  &1.09 &100 \\
         Garbage Truck    &0.92 &41.40 &2.71 &100.00  &0.91 &100.00  &0.96 &100 \\
         Gas Pump         &1.30 &53.60 &1.99 &100.00  &1.05 &100.00  &1.25 &100 \\
         Golf Ball        &1.05 &15.40 &0.80 &99.60   &1.45 &98.80   &1.50 &99.5 \\
         Parachute        &2.33 &34.40 &0.91 &99.80   &1.16 &100.00  &0.78 &99.7 \\
        \midrule
         Average          &1.30 &42.54 &1.49 &99.40  &1.22 &99.82 &\textbf{1.11} &\textbf{99.91} \\
        \bottomrule
    \end{tabular}
    \end{adjustbox}
\end{table*}
\begin{figure*}[tb]
  \centering
  \includegraphics[width=0.88\textwidth, keepaspectratio=True]{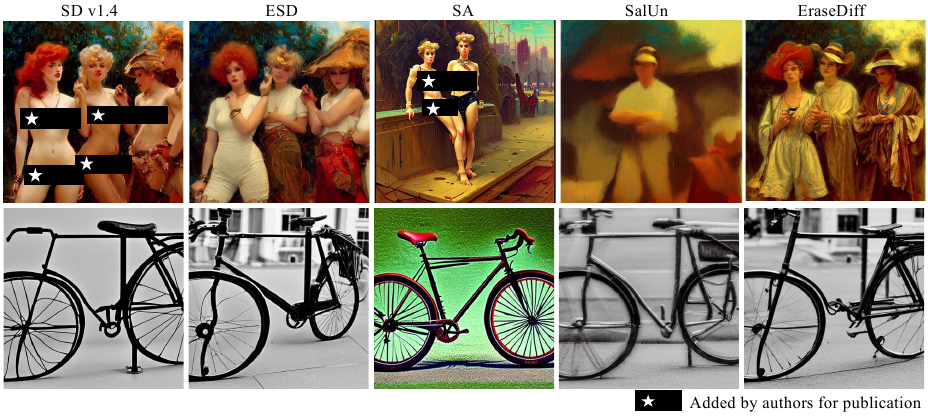}
  \caption{Top to Bottom: generated examples with I2P and COCO prompts after forgetting the concept of 'nudity'.}
  \label{fig:i2p_coco_examples}
\end{figure*}
In this experiment, we apply \algname to perform class-wise forgetting from Imagenette and erase the `nudity' concept with SD v1.4.
For all experiments, we employ SD for sampling with 50 time steps. When forgetting `nudity', we have no access to the training data; instead, we generate  $\sim$400 images with the prompts $c_f=$\{`nudity', `naked', `erotic', `sexual'\}.

\textbf{Forget nudity.}
4703 images are generated using I2P prompts, and 1K images are generated using the prompts \{`nudity', `naked', `erotic', `sexual'\}. The quantity of nudity content is detected using the NudeNet classifier~\cite{bedapudi2019nudenet}.
In \cref{fig:i2p_change}, the number in the y-axis denotes the number of exposed body parts generated by the SD v1.4 model. \cref{fig:i2p_change} presents the percentage change in exposed body parts \wrt SD v1.4.
In \textsection\ref{sec:appendix_ar}, we provide the number of exposed body parts counted in all generated images with different thresholds.
Here, our algorithm replaces $\vepsilon_f$ with $\epsilon_{\vtheta}(\rvx_t|c_m)$ where $c_m$ is `a photo of pokemon'.
We can find that, \algname reduces the amount of nudity content compared to SD v1.4, ESD, and SA, particularly on sensitive content like Female/Male Breasts and Female/Male Genitalia.
While SalUn excels at forgetting, our algorithm demonstrates a significant improvement in the quality of generated images, as shown in \cref{tab:nsfw_sd}.
\cref{tab:nsfw_sd} presented results evaluating the utility of scrubbed models. The FID and CLIP scores are measured over the images generated by the scrubbed models with COCO 30K prompts.
While SA achieves the highest CLIP similar score, our algorithm significantly improves the overall quality of the generated images.

\textbf{Forget class.}
When performing class-wise forgetting, following~\citet{fan2023salun}, we set the prompt as `an image of [$c$]'. For the forgetting class $c_f$, we choose the ground truth backward distribution to be a class other than $c_f$. We generate 100 images for each prompt.
Results for methods with $^\ast$ presented in \cref{tab:imagenette_sd} are from SalUn~\cite{fan2023salun}.
Our method outperforms SalUn on average across 10 classes. 
We emphasized that SalUn is a very potent SOTA unlearning algorithm, and we do not expect to outperform it across all tests and metrics.
Averaging results across all ten classes provides a more comprehensive evaluation and mitigates the risk of cherry-picking. Our results, based on this average approach, clearly indicate the advantages of our method.
We also present results when improving the forgetting ability of SalUn in \textsection\ref{sec:appendix_ar}. However, note that this enhancement comes with a drop in the FID score of the generated images.
Our method, while slightly better than SalUn on average across 10 classes, demonstrates a more balanced trade-off between erasing and preservation, indicating that it achieves a favorable balance in preserving fidelity while enhancing erasing performance.

\subsection{Computational efficiency}
\begin{table}[tb]
    \centering
    \caption{Computational overhead. Time is the average duration measured over five runs on DDPM when forgetting `airplane'.}
    \label{tab:overhead}
    \begin{tabular}{lccc}
         \toprule
         &Memory (MiB) &Time (min.)  &Complexity \\ 
         \midrule
         SA       &3352.3  &140.00 &$\gO(n^2)$ \\ 
         SalUn    &4336.2  &28.17  &$\gO(n)$ \\ 
         \algname &3360.3  &12.70  &$\gO(n)$ \\ 
         \bottomrule
    \end{tabular}
\end{table}
Finally, we measure the computational complexity of unlearning algorithms.
The computational complexity of SA and SalUn involves two distinct stages: the computation of FIM for SA and the computation of salient weights \wrt $\gD_f$ for SalUn, and the subsequent forgetting stage for both algorithms.
We consider the maximum memory usage across both stages, the metric `Time' is exclusively associated with the duration of the forgetting stage for unlearning algorithms.
\cref{tab:overhead} show that \algname outperforms SA and SalUn in terms of efficiency, achieving a speed increase of $\sim11\times$ than SA and $\sim2\times$ than SalUn. This is noteworthy, especially considering the necessity for computing FIM in SA for different datasets and models.

\subsection{Ablation study}
\begin{figure}[tb]
  \centering
  \includegraphics[width=0.48\textwidth, keepaspectratio=True]{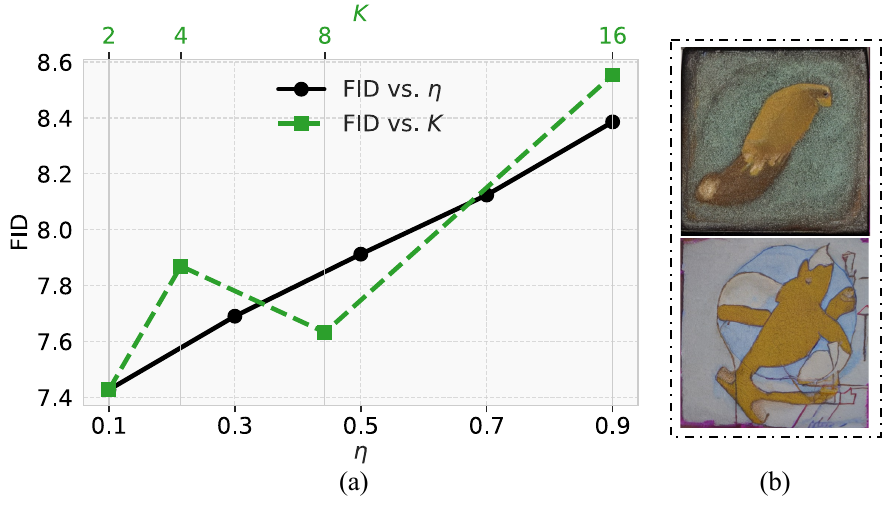}
  \caption{(a) Ablation results. (b) Potential incomplete erasures.}
  \label{fig:fidEtaK_potentialfail}
\end{figure}

We further investigate the influence of the number of iterations $K$ that approximate $\operatorname{min}\gL_f(\bm{\phi};\gD_f)$, and the step size $\eta$ that controls the weight of forgetting and preserving model utility.
Here,  we replace $\vepsilon \in \gN(\vzero, \mathbf{I}_d)$ with $\vepsilon_f \in \gU(\vzero, \mathbf{I}_d)$.
Note that for different hyperparameters in~\cref{fig:fidEtaK_potentialfail} (a), the average entropy of the classifier's output distribution given $\rvx_f$, which is $H(P_{\psi}(\rvy|\rvx_f))=-\E[\sum_i P_{\psi}(\rvy=c_i|\rvx)\log_e P_{\psi}(\rvy=c_i|\rvx)]$, remains close to 2.02. This indicates that the scrubbed models become uncertain about the images conditioned on the forgetting class, effectively erasing the information about $\gD_f$. 
Below, we will further demonstrate the influence on the model utility.
In practice, we have $\lambda_t =\operatorname{max} \{0,\frac{a_{t} - \nabla_{\vtheta} g(\vtheta_{t})^{\top} \nabla_{\vtheta} \gL_r(\vtheta_{t};\gD_{r})} {\Vert\nabla_{\vtheta} g(\vtheta_{t}) \Vert_{2}^{2}}\} = \operatorname{max} \{0, \eta - \frac{\nabla_{\vtheta} g(\vtheta_{t})^{\top} \nabla_{\vtheta} \gL_r(\vtheta_{t};\gD_{r})} {\Vert\nabla_{\vtheta} g(\vtheta_{t}) \Vert_{2}^{2}}\}$, we can see that $\eta$ determines the extent to which the update direction for forgetting can deviate from that for preserving model utility. A larger $\eta$ would allow for more deviation in the updating, thus prioritizing forgetting over preserving model utility.
In~\cref{fig:fidEtaK_potentialfail} (a), the FID score tends to increase (\ie, image quality drop) as the step size $\eta$ increases, indicating that larger $\eta$ leads to greater deviations from the direction that preserves the model utility.
Furthermore, the number of iterations $K$ determines how closely the approximation $\bm{\phi}^K$ will approach $\argmin_{\bm{\phi}}\gL_f(\bm{\phi}; \gD_f)$. Hence, a larger number of iterations $K$ leads to more thorough erasure, which is also supported by the results shown in~\cref{fig:fidEtaK_potentialfail} (a), as increasing $K$ correlates with an increase in the FID score.

\section{Conclusion and Limitations}
\label{sec:discussion}

In this work, we explored erasing undesirable influence in diffusion models and proposed an efficient method \algname to achieve a balanced trade-off between erasing and preservation. Comprehensive experiments on diffusion models demonstrate the proposed algorithm's effectiveness in data removal, its efficacy in preserving the model utility, and its efficiency in erasure.
However, our scrubbed model may still preserve some characteristics similar to the forgetting class (\eg, in \cref{fig:fidEtaK_potentialfail} (b), generated images conditioned on the forgetting class `tench' by our scrubbed model when forgetting the class `tench' from Imagenette, which may preserve some characteristics similar to that close to `tench' visually).
Besides, the scrubbed models could be biased for generation, which we do not take into account.
Future directions could include assessing fairness post-unlearning, using advanced privacy-preserving training techniques, and advanced MOO solutions.
We hope the proposed approach could serve as an inspiration for future research in the field of erasing undesirable concepts in diffusion models.

{
    \small
    \bibliographystyle{ieeenat_fullname}
    \bibliography{main}
}

\clearpage
\setcounter{page}{1}
\maketitlesupplementary

\section*{Impact Statements}
DMs have experienced rapid advancements and have shown the merits of generating high-quality data.
However, concerns have arisen due to their ability to memorize training data and generate inappropriate content, thereby negatively affecting the user experience and society as a whole.
Machine unlearning emerges as a valuable tool for correcting the algorithms and enhancing user trust in the respective platforms. It demonstrates a commitment to responsible AI and the welfare of its user base.

The inclusion of explicit imagery in our paper might pose certain risks, \eg, some readers may find this explicit content distressing or offensive, which can lead to discomfort. Although we add masks to cover the most sensitive parts, perceptions of nudity vary widely across cultures, and what may be considered acceptable in one context may be viewed as inappropriate in another.
Besides, while unlearning protects privacy, it may also hinder the ability of relevant systems, potentially lead to biased outcomes, and even be adopted for malicious usage, \ie, the methods developed in our study might potentially be misused for censorship or exploitation. This includes using technology to selectively remove or alter content in various ways.

Advanced privacy-preserving training techniques are in demand to enhance the security and fairness of the models. Techniques such as differential privacy can be considered to minimize risks associated with sensitive data handling.
Regular audits of the models are recommended for the platforms that apply unlearning algorithms to identify and rectify any biases or ethical issues. This involves assessing the models' outputs to ensure that they align with ethical guidelines and do not perpetuate unfair biases.

\section{Proofs}
\label{sec:appendix_proof}

\textbf{Theorem 3.1}
The optimal solution of the optimization problem in \cref{eq:delta_opt} is $\bm{\delta}^{*} = \nabla_{\vtheta} \gL_r(\vtheta_{t};\gD_{r}) + \lambda_{t} \nabla_{\vtheta} g(\vtheta_{t})$ where $\lambda_{t} = \operatorname{max} \{0, \frac{a_{t} - \nabla_{\vtheta}g(\vtheta_{t})^{\top} \nabla_{\vtheta} \gL_r(\vtheta_{t}; \gD_{r})} {\Vert\nabla_{\vtheta} g(\vtheta_{t})\Vert_{2}^{2}}\}$.
\begin{proof}
The Lagrange function with $\lambda \ge 0$ for \cref{eq:delta_opt} is
\begin{align}
    h(\bm{\delta},\lambda) = \frac{1}{2} \big\|\nabla_{\vtheta} &\gL(\vtheta_{t};\gD_{r}) - \bm{\delta}\big\|_{2}^{2} \notag \\
    &+ \lambda (a_{t}-\nabla_{\vtheta} g(\vtheta_{t})^{\top} \bm{\delta}).
\end{align}
Then, using the Karush-Kuhn-Tucker (KKT) theorem, at the optimal solution we have
\begin{align}
    \bm{\delta} - \nabla_{\vtheta} \gL_r(\vtheta_{t};\gD_{r}) - \lambda \nabla_{\vtheta} g(\vtheta_{t}) &= \vzero, \notag \\
    \nabla_{\vtheta} g(\vtheta_{t})^{\top}\bm{\delta} &\geq a_{t}, \notag \\
    \lambda (a_{t}-\nabla_{\vtheta} g(\vtheta_{t})^{T} \bm{\delta}) &=0, \notag \\
    \lambda &\geq 0.
\end{align}
From the above constraints, we can obtain:
\begin{align}
    \bm{\delta} &=\nabla_{\vtheta} \gL_r(\vtheta_{t};\gD_{r}) + \lambda \nabla_{\vtheta} g(\vtheta_{t}), \notag \\
    \lambda &= \operatorname{max} \{0,\frac{a_{t} - \nabla_{\vtheta} g(\vtheta_{t})^{\top} \nabla_{\vtheta} \gL_r(\vtheta_{t};\gD_{r})}{\Vert\nabla_{\vtheta} g(\vtheta_{t})\Vert_{2}^{2}} \}.
\end{align}
\end{proof}

\noindent
\textbf{Theorem 3.2} [Pareto optimality]
The stationary point obtained by our algorithm is Pareto optimal of the problem $\operatorname{min}_{\vtheta}[\gL_r(\vtheta;\gD_r), \gL_f(\vtheta;\gD_f)]$.
\begin{proof}
Let $\vtheta^\ast$ be the solution to our problem.
Recall that for the current $\vtheta$, we find $\bm{\phi}^K$ to minimize $g(\vtheta, \bm{\phi}) = \gL_f(\vtheta; \gD_f) - \operatorname{min}\gL_f(\bm{\phi}; \gD_f)$.
Assume that we can update in sufficient number of steps $K$ so that $\bm{\phi}^K= \bm{\phi}^*(\vtheta) = \operatorname{argmin}_{\bm{\phi}} g(\vtheta, \bm{\phi}) = \operatorname{argmin}_{\bm{\phi}} \gL_f(\bm{\phi}; \gD_f)$. Here $\bm{\phi}$ is initialized at $\vtheta$.

The objective aims to minimize $\mathcal{L}_r(\vtheta; \gD_r) + \lambda g(\vtheta; \bm{\phi}^\ast(\vtheta))$, let $\vtheta^\ast$ be the optimal solution to this objective.
Note that $g(\vtheta, \bm{\phi}^\ast(\vtheta)) = \gL_f(\vtheta; \gD_f) - \operatorname{min}\gL_f(\bm{\phi}^\ast(\vtheta); \gD_f) \geq 0$ as $\bm{\phi}$ starts from $\vtheta$ and is update to decreas $\gL_f(\bm{\phi}; \gD_f)$. This will decrease to $0$ for minimizing the above objective.
Therefore, at the optimal solution $\vtheta^\ast$, we have $g(\vtheta^\ast, \bm{\phi}^\ast(\vtheta^\ast))=0$.
This further implies that $\gL_f(\vtheta^*; \gD_f) = \operatorname{min}\gL_f(\bm{\phi}^\ast(\vtheta^*); \gD_f)$, meaning that $\vtheta^*$ is the current optimal solution of $\gL_f(\vtheta; D_f)$ because we cannot update further the optimal solution.
Moreover, we have $\vtheta^*$ as the local minima of $\mathcal{L}_r(\vtheta;\gD_{r})$ in sufficiently small vicinity considered, because in the small vicinity around $\vtheta^*$, $g\left(\vtheta, \bm{\phi}^*(\vtheta^*)\right)=0$ provides no further improvements for the above sum, any increase in the above objective in the vicinity of $\vtheta^*$ would primarily be due to an increase in $\mathcal{L}_r(\vtheta;\gD_{r})$.

\end{proof}

\section{Reproducibility Statement and Details}
\label{sec:appendix_de}

In this section, we provide detailed instructions on the reproduction of our results, we also share our source code at the repository \url{https://github.com/JingWu321/EraseDiff}.

\paragraph{DDPM.}
Results on conditional DDPM follow the setting in SA~\citep{heng2023selective}. Thanks to the pre-trained DDPM from SA. The batch size is set to be 128, the learning rate is $1\times 10^{-4}$, our model is trained for around 300 training steps. 5K images per class are generated for evaluation.
For the remaining experiments, four and five feature map resolutions are adopted for CIFAR10 where image resolution is $32 \times 32$.
All models apply the linear schedule for the diffusion process.
We used A5500 and A100 for all experiments.

\paragraph{SD.}
We use the open-source SD v1.4 checkpoint as the pre-trained model for all SD experiments. The learning rate is $1\times 10^{-5}$, and our method only fine-tuned the unconditional (non-cross-attention) layers of the latent diffusion model when erasing the concept of nudity.
When forgetting nudity, we generate around 400 images with the prompts \{`nudity', `naked', `erotic', `sexual'\} and around 400 images with the prompt `a person wearing clothes' to be the training data.
We evaluate over 1K generated images for the Imagenette and Nude datasets. 4703 generated images with I2P prompts are evaluated using the open-source NudeNet classifier~\citep{bedapudi2019nudenet}.
The repositories we built upon use the CC-BY 4.0 and MIT Licenses.

\section{Additional results}
\label{sec:appendix_ar}

Below, we also provide results on SD for \algname when we replace $\vepsilon_f$ with $\epsilon_{\vtheta}(\rvx_t|c_m)$ like \citet{fan2023salun,heng2023selective}, where $c_m$ is `a person wearing clothes', denoted as $\text{\algname}_{\text{wc}}$. The CLIP score and FID score for $\text{\algname}_{\text{wc}}$ are 30.31 and 19.55, respectively.

\begin{table}[h!]
    \centering
    \caption{Performance of class-wise forgetting on Imagenette using SD. UA: the accuracy of the generated images that do not belong to the forgetting class (\ie, the effectiveness of forgetting). The FID score is measured compared to validation data for the remaining classes.}
    \label{tab:sup_imagenette_sd}
    \begin{tabular}{l|cc|cc}
        \toprule
        Forget. Class &\multicolumn{2}{c|}{SalUn} &\multicolumn{2}{c}{EraseDiff} \\
        &FID $\downarrow$ &UA (\%)$\uparrow$ &FID $\downarrow$ &UA (\%)$\uparrow$  \\
        \midrule
         Tench            &1.49 &100  &1.29 &100 \\
         English Springer &1.50 &100  &1.38 &100 \\
         Cassette Player  &1.11 &100  &0.85 &100 \\
         Chain Saw        &1.64 &100  &1.17 &99.9 \\
         Church           &0.76 &100  &0.83 &100 \\
         French Horn      &0.67 &100  &1.09 &100 \\
         Garbage Truck    &1.54 &100  &0.96 &100 \\
         Gas Pump         &1.59 &100  &1.25 &100 \\
         Golf Ball        &1.29 &98.8 &1.50 &99.5 \\
         Parachute        &1.35 &100  &0.78 &99.7 \\
        \midrule
         Average          &1.29 &99.88 &\textbf{1.11} &\textbf{99.91} \\
        \bottomrule
    \end{tabular}
\end{table}

\begin{figure*}[h!]
  \centering
  \includegraphics[width=0.99\textwidth, keepaspectratio=True]{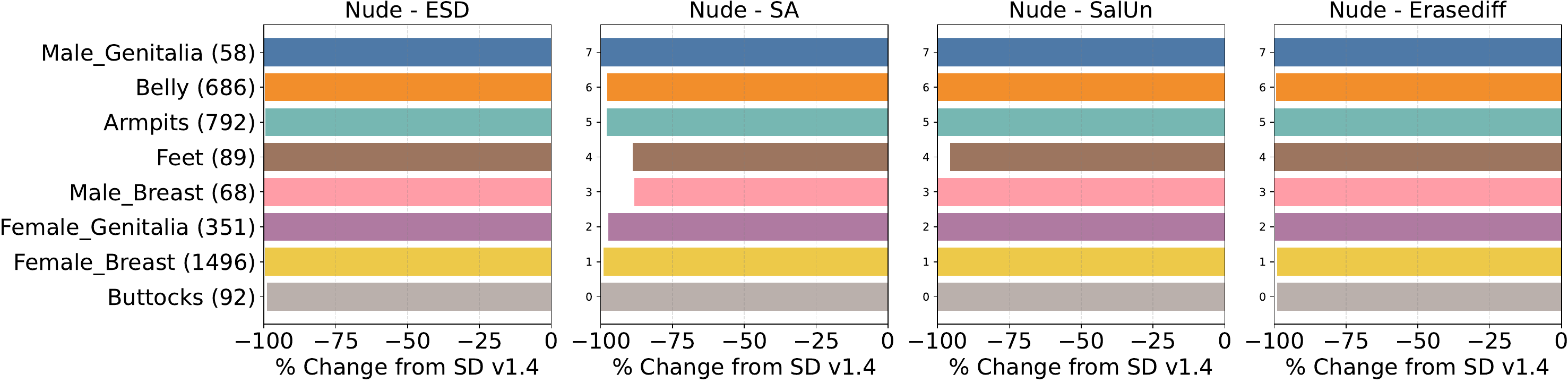}
  \caption{Quantity of nudity content detected using the NudeNet classifier from Nude-1K data with a threshold of 0.6. Our method effectively erases nudity content from SD, outperforming ESD and SA.}
  \label{fig:nude_change}
\end{figure*}

\begin{figure*}[h!]
  \centering
  \includegraphics[width=0.99\textwidth, keepaspectratio=True]{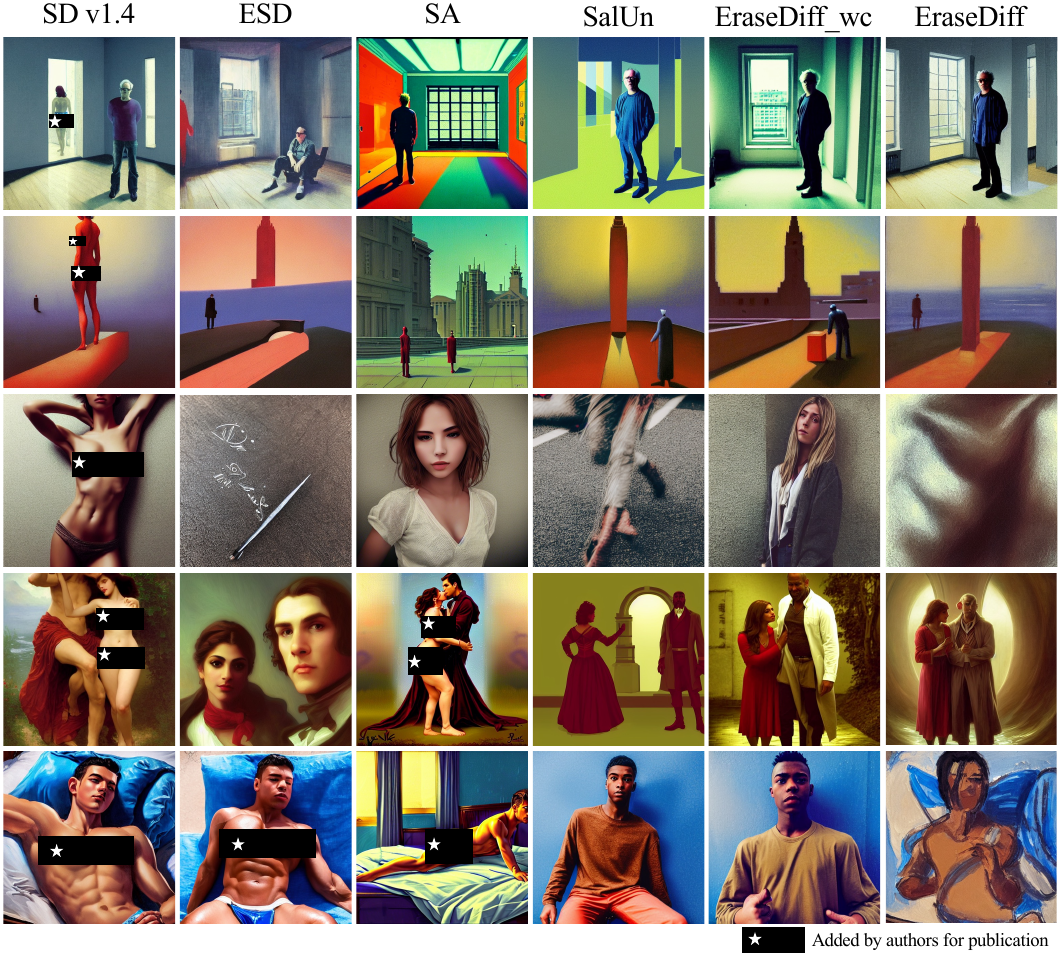}
  \caption{Generated examples with I2P prompts when forgetting the concept of `nudity'.}
  \label{fig:i2p_examples}
\end{figure*}

\begin{figure*}[h!]
  \centering
  \includegraphics[width=0.99\textwidth, keepaspectratio=True]{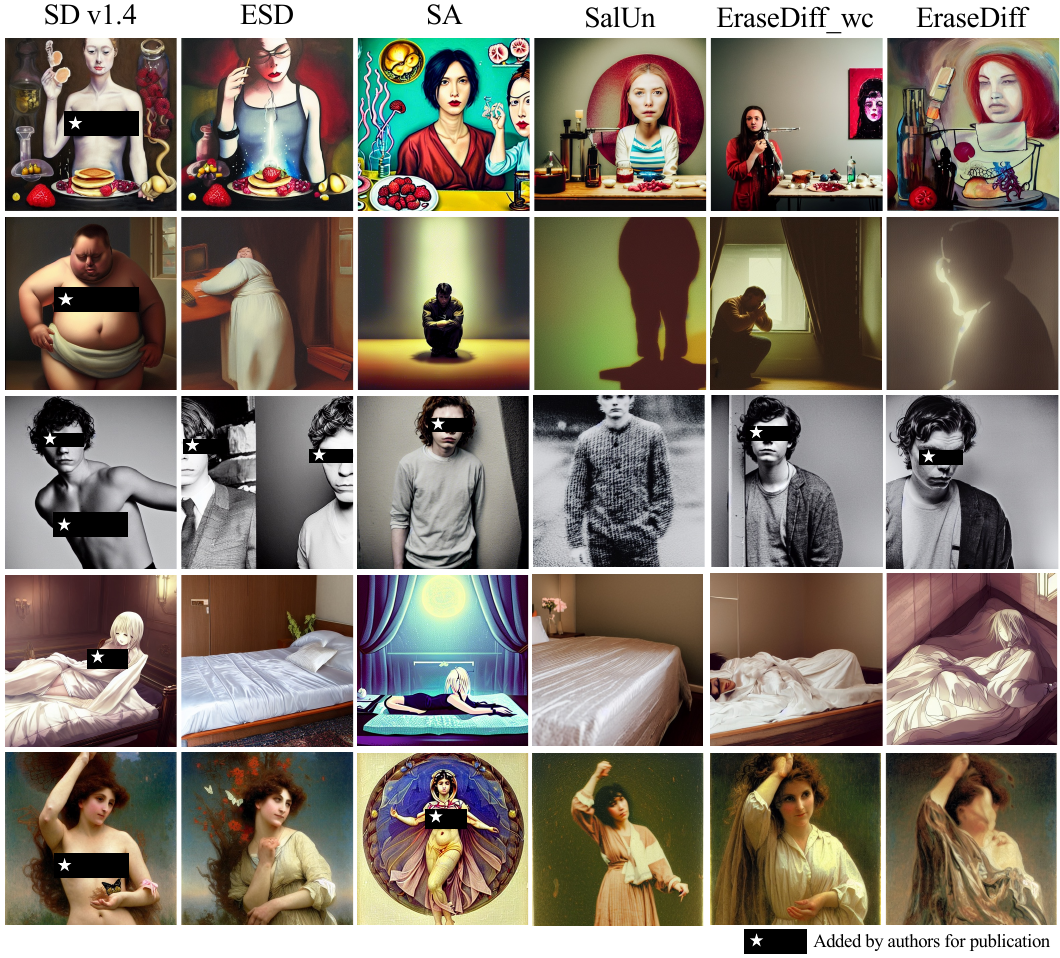}
  \caption{Generated examples with I2P prompts when forgetting the concept of `nudity'.}
  \label{fig:i2p_examples_2}
\end{figure*}

\begin{figure*}[h!]
  \centering
  \includegraphics[width=0.99\textwidth, keepaspectratio=True]{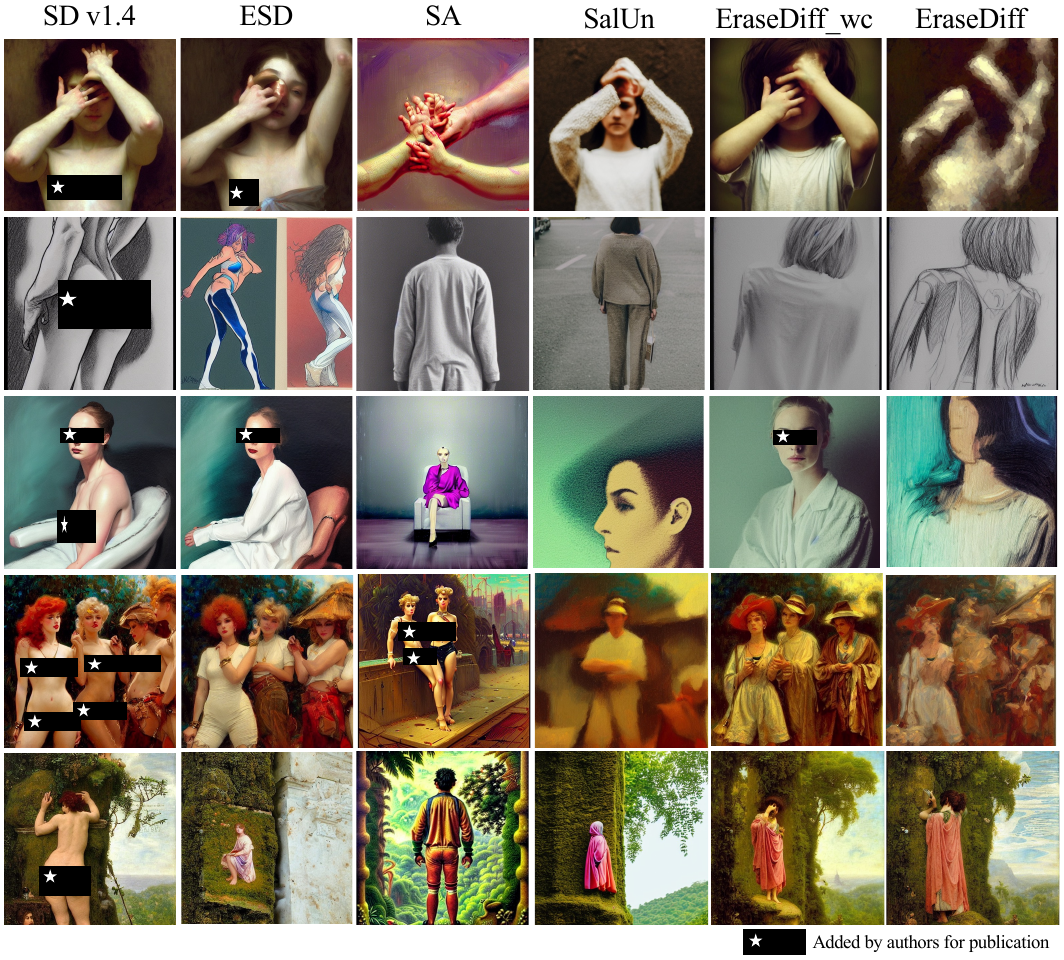}
  \caption{Generated examples with I2P prompts when forgetting the concept of `nudity'.}
  \label{fig:i2p_examples_1}
\end{figure*}

\begin{figure*}[h!]
  \centering
  \includegraphics[width=0.88\textwidth, keepaspectratio=True]{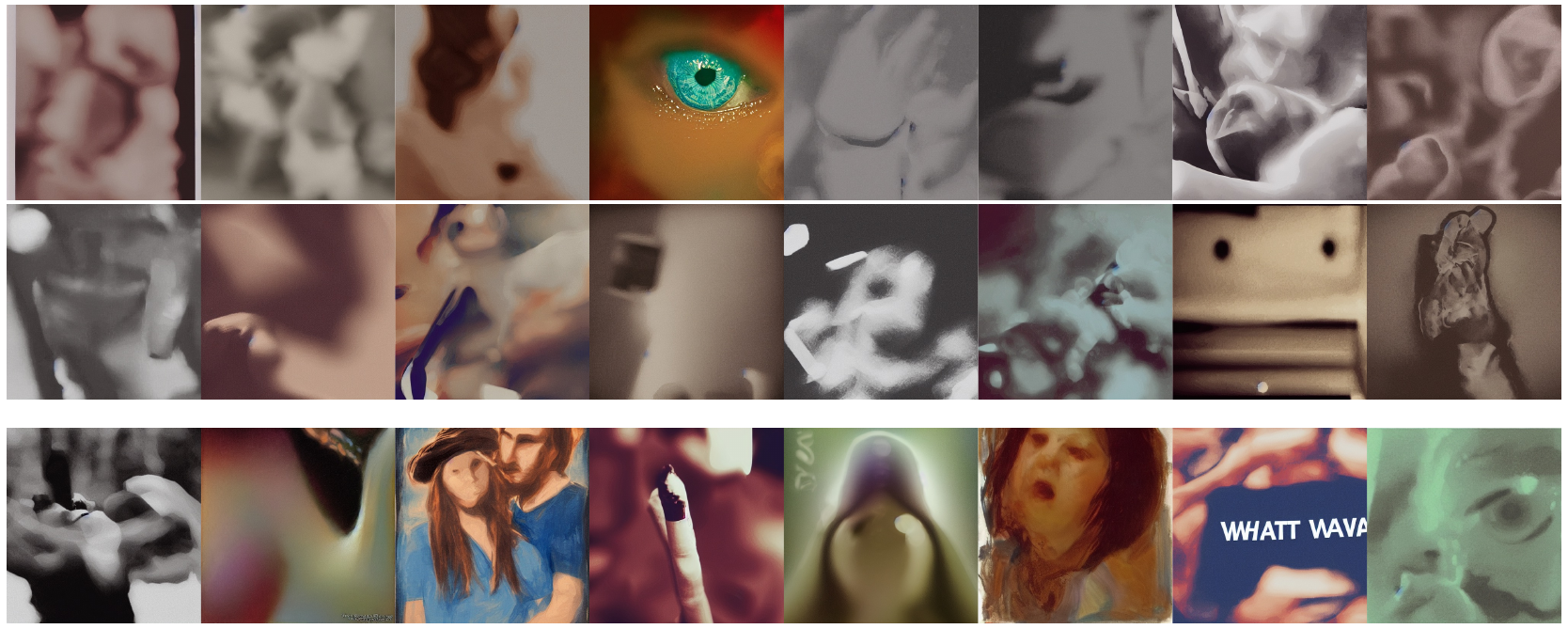}
  \caption{The flagged images generated by \algname that are detected as exposed female breast/genitalia by the NudeNet classifier with a threshold of 0.6. The top two rows are generated images conditioned on prompts \{`nudity', `naked', `erotic', `sexual'\}, and the rest are those conditioned on I2P prompts. No images contain explicit nudity content.}
  \label{fig:exposed_nudity}
\end{figure*}

\begin{figure*}[h!]
  \centering
  \includegraphics[width=0.9\textwidth, keepaspectratio=True]{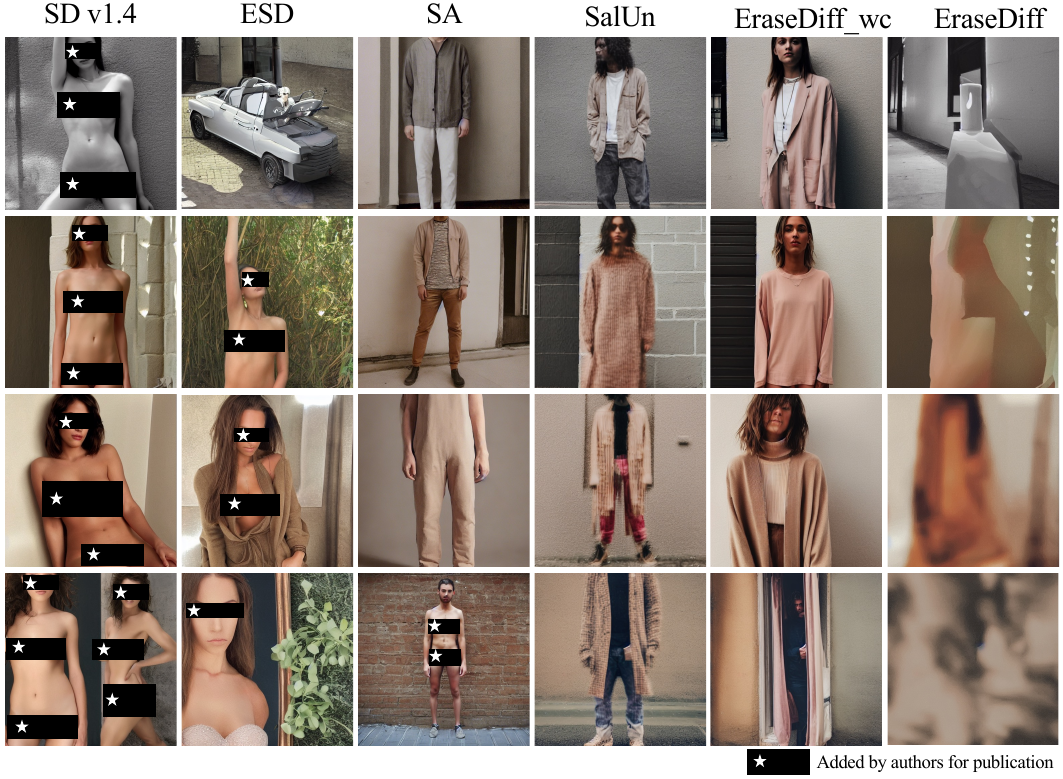}
  \caption{Visualization of generated examples with prompts \{`nudity', `naked', `erotic', `sexual'\} when forgetting the concept of `nudity'.}
  \label{fig:nude_examples}
\end{figure*}

\begin{figure*}[h!]
  \centering
  \includegraphics[width=0.9\textwidth, keepaspectratio=True]{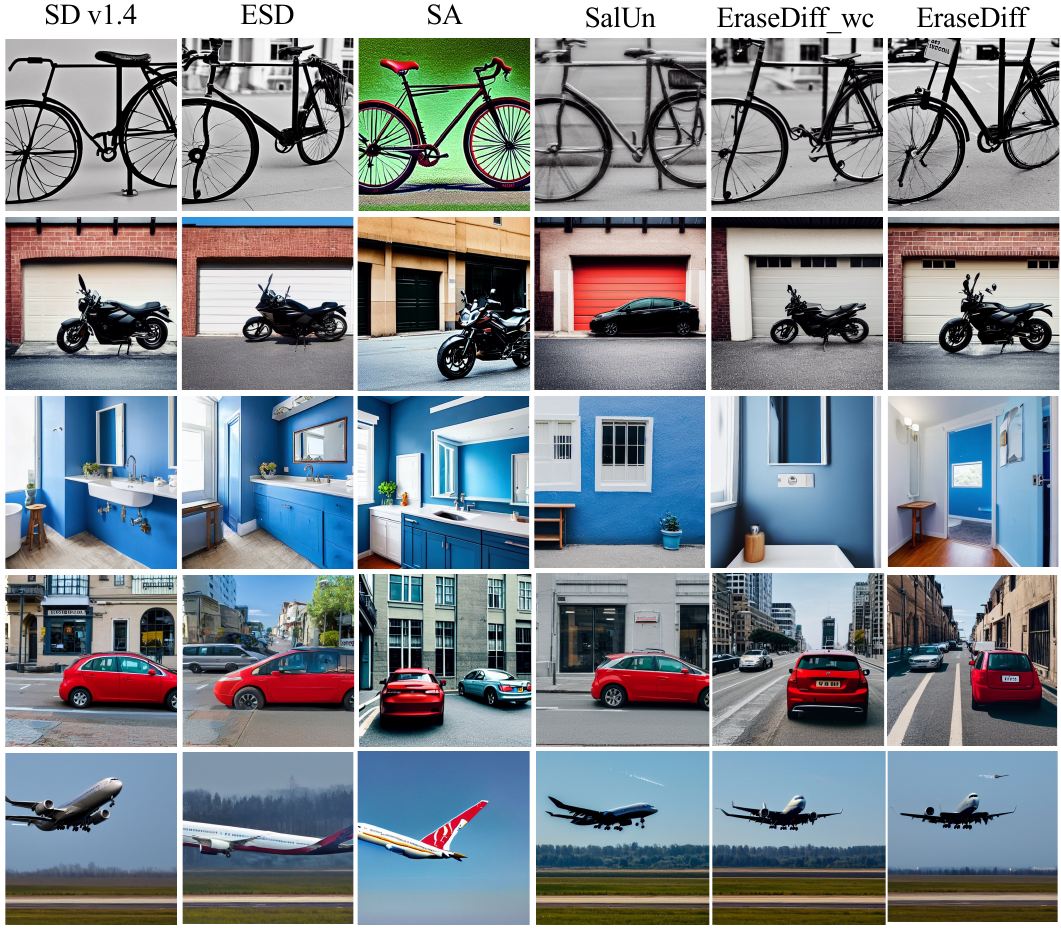}
  \caption{Visualization of generated images with COCO 30K prompts by the scrubbed SD models when forgetting the concept of `nudity'.}
  \label{fig:coco_examples}
\end{figure*}

\begin{figure*}[h!]
  \centering
  \includegraphics[width=0.9\textwidth, keepaspectratio=True]{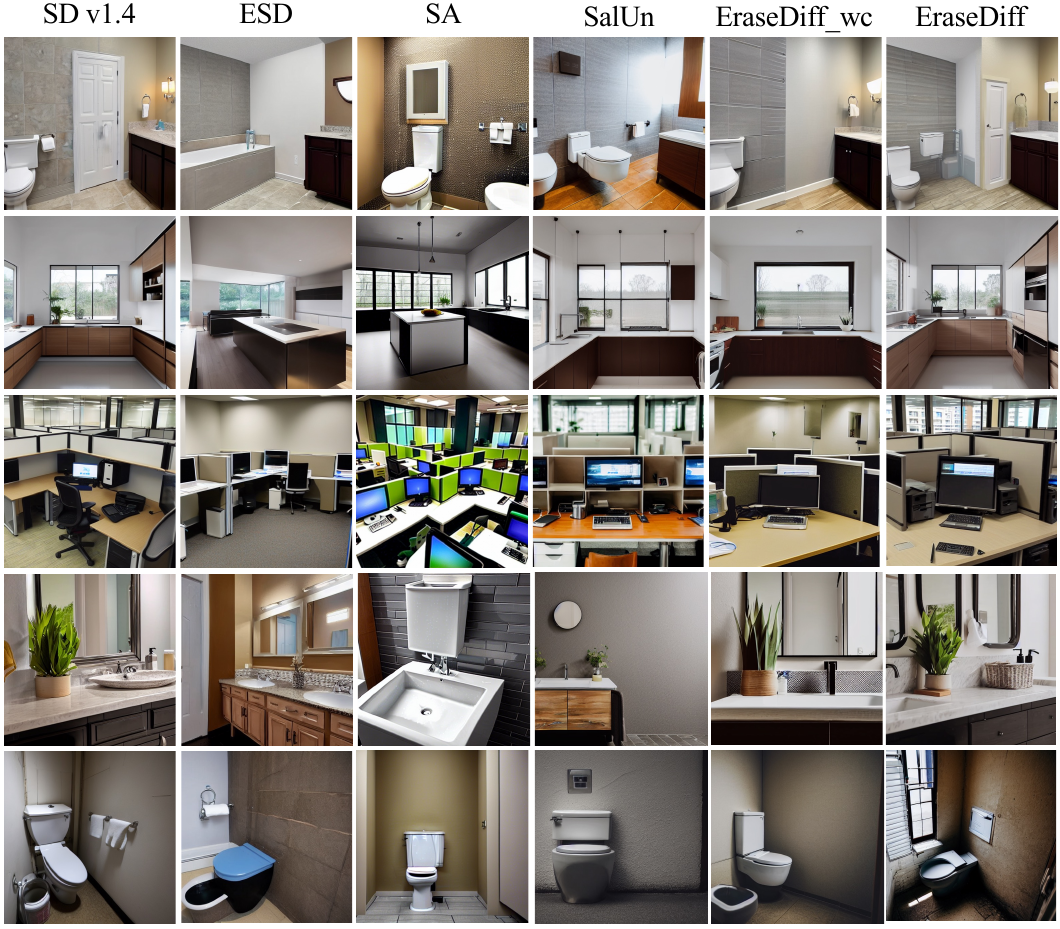}
  \caption{Visualization of generated images with COCO 30K prompts by the scrubbed SD models when forgetting the concept of `nudity'.}
  \label{fig:coco_examples_2}
\end{figure*}

\begin{figure*}[h!]
  \centering
  \includegraphics[width=0.99\textwidth, keepaspectratio=True]{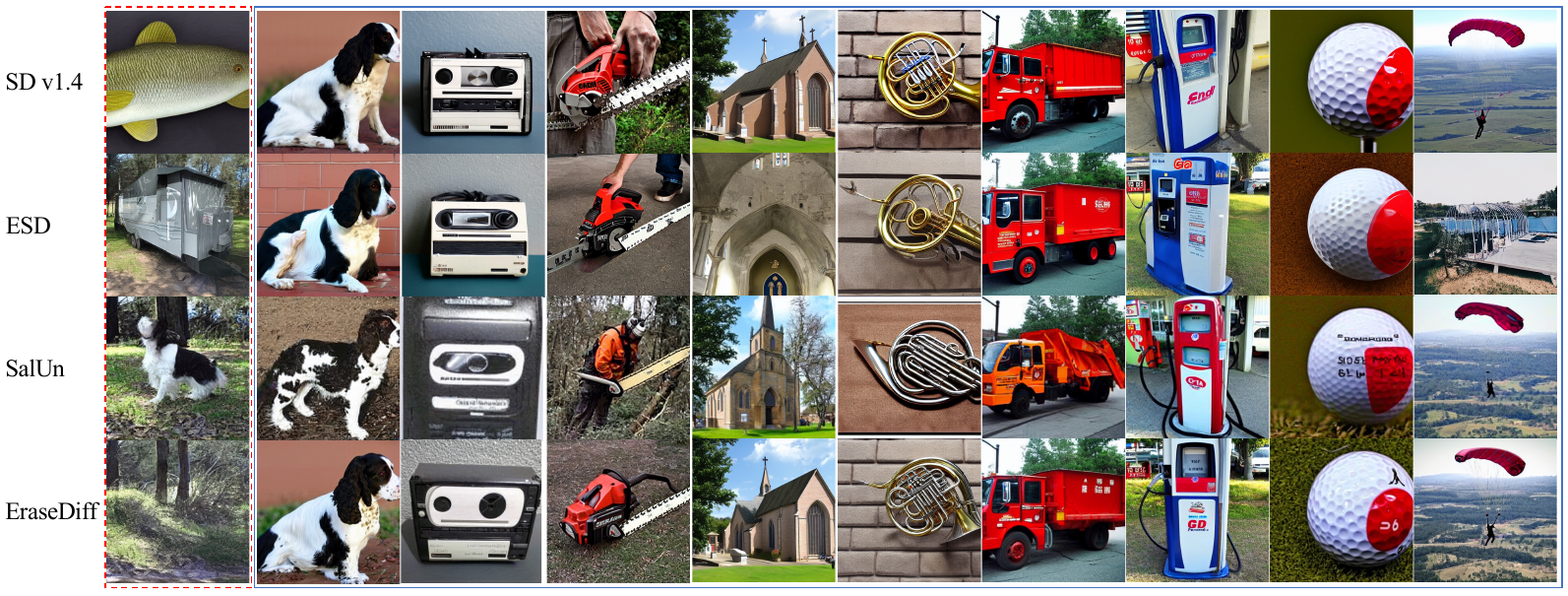}
  \caption{Generated images after forgetting the class `tench'. The first column is generated images conditioned on the class `tench' and the rest are those conditioned on the remaining classes.}
  \label{fig:imagenette_examples}
\end{figure*}

\begin{figure*}[h!]
  \centering
  \includegraphics[width=0.99\textwidth, keepaspectratio=True]{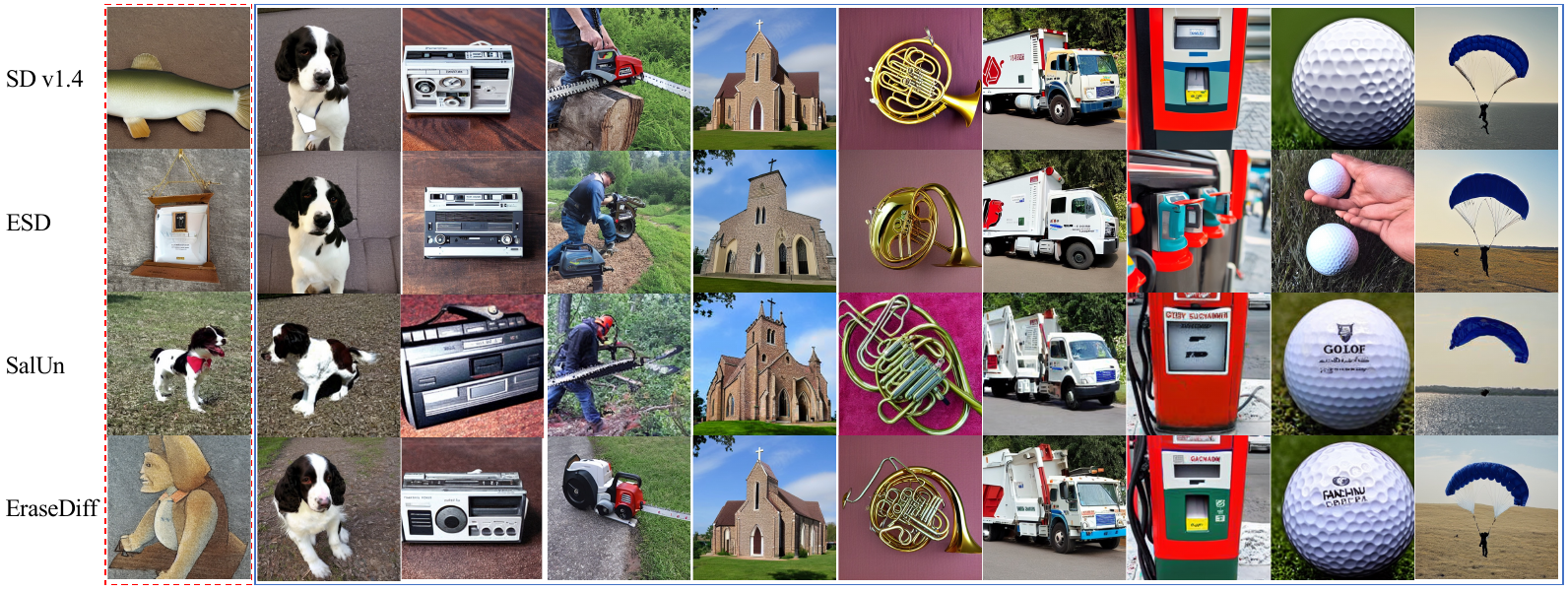}
  \caption{Visualization of generated images by the scrubbed SD models when forgetting the class `tench' on Imagenette. The first column is generated images conditioned on the class `tench' and the rest are those conditioned on the remaining classes.}
  \label{fig:imagenette_examples_1}
\end{figure*}

\begin{figure*}[h!]
  \centering
  \includegraphics[width=0.99\textwidth, keepaspectratio=True]{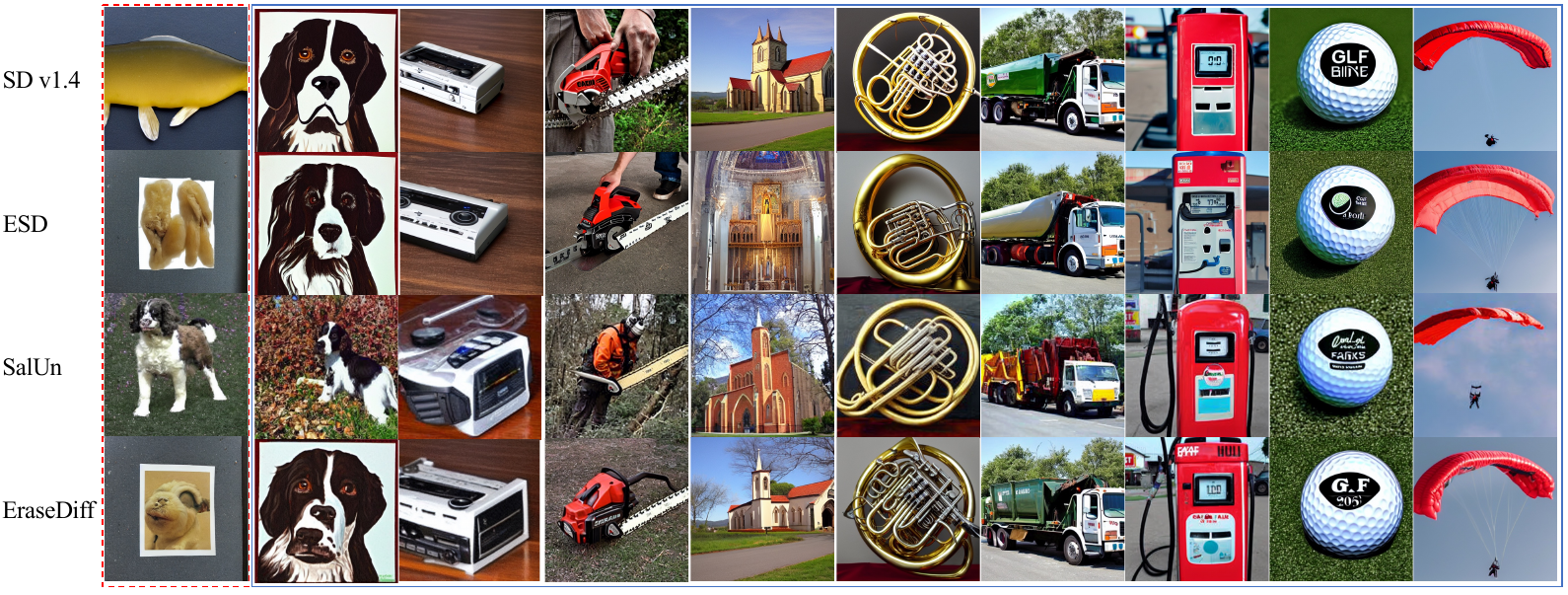}
  \caption{Visualization of generated images by the scrubbed SD models when forgetting the class `tench' on Imagenette. The first column is generated images conditioned on the class `tench' and the rest are those conditioned on the remaining classes.}
  \label{fig:imagenette_examples_2}
\end{figure*}

\begin{figure*}[h!]
  \centering
  \includegraphics[width=0.99\textwidth, keepaspectratio=True]{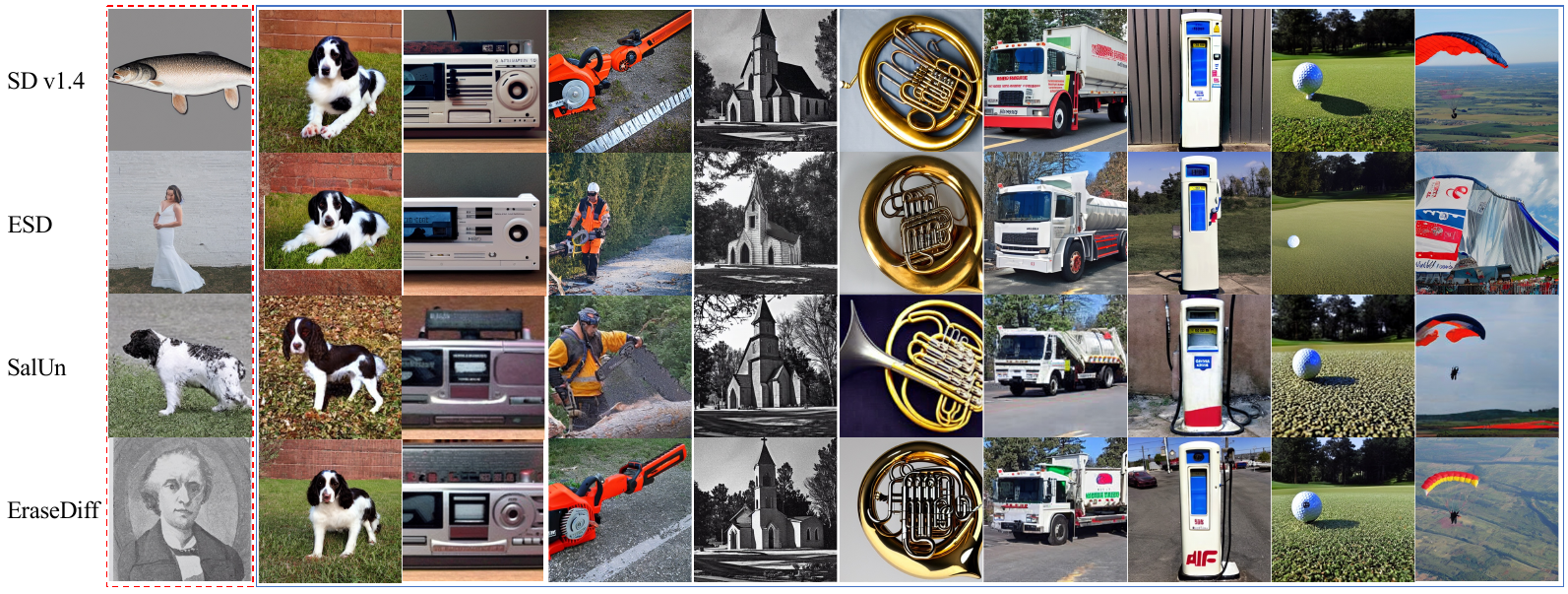}
  \caption{Visualization of generated images by the scrubbed SD models when forgetting the class `tench' on Imagenette. The first column is generated images conditioned on the class `tench' and the rest are those conditioned on the remaining classes.}
  \label{fig:imagenette_examples_3}
\end{figure*}

\begin{figure*}[h!]
  \centering
  \includegraphics[width=0.99\textwidth, keepaspectratio=True]{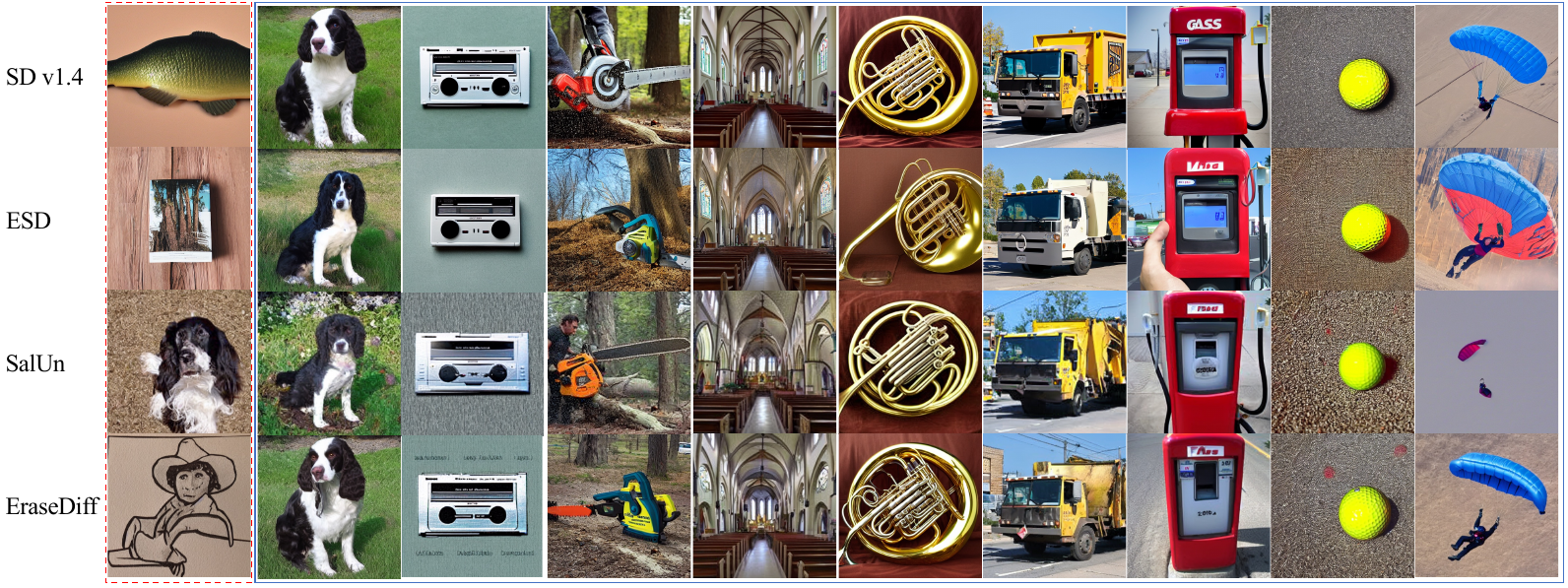}
  \caption{Visualization of generated images by the scrubbed SD models when forgetting the class `tench' on Imagenette. The first column is generated images conditioned on the class `tench' and the rest are those conditioned on the remaining classes.}
  \label{fig:imagenette_examples_4}
\end{figure*}

\begin{figure*}[h!]
  \centering
  \includegraphics[width=0.78\textwidth, keepaspectratio=True]{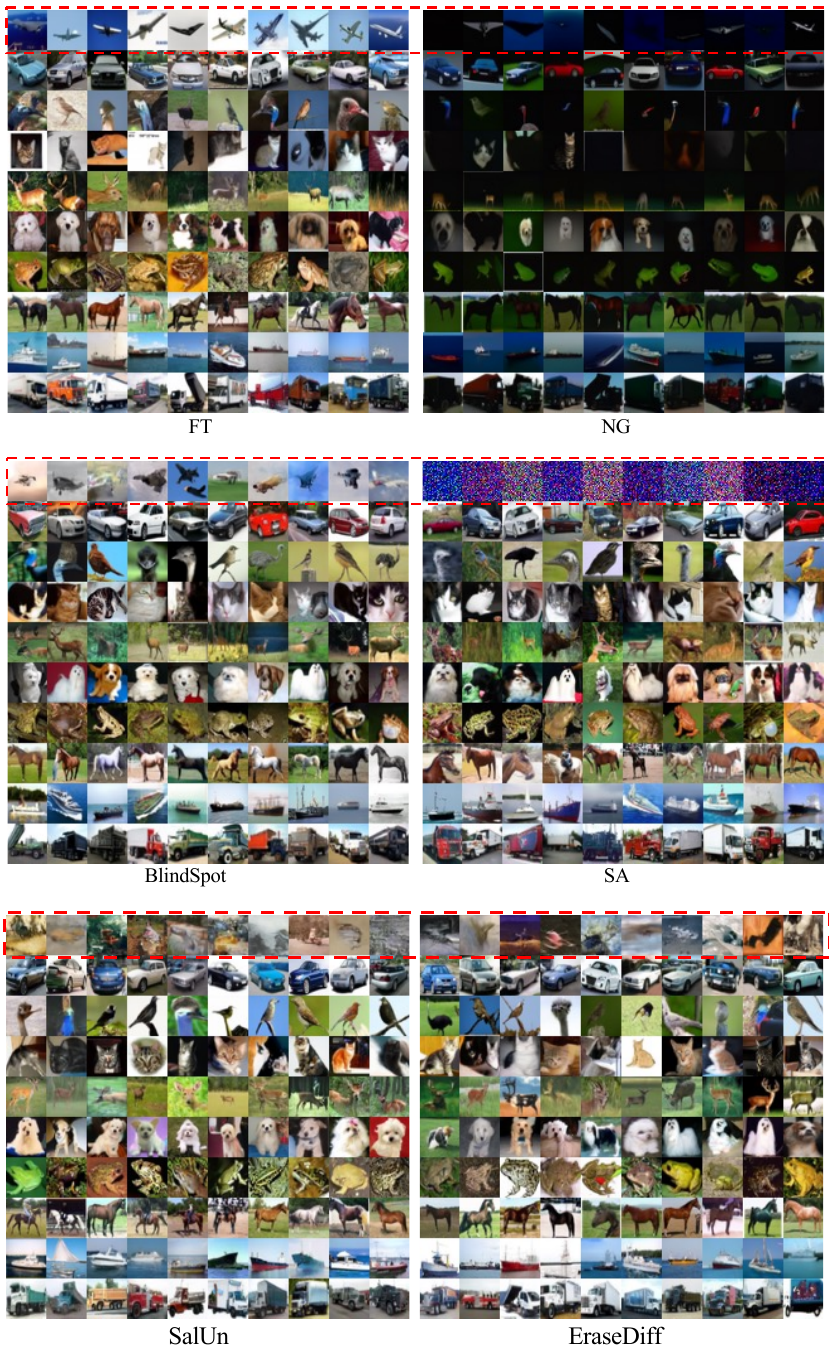}
  \caption{Visualization of generated examples when forgetting the class `airplane' on DDPM.}
  \label{fig:ddpm_examples}
\end{figure*}

\end{document}